\documentclass[10pt,twocolumn,letterpaper]{article}

\usepackage[algorithms,algorithms]{wacv}      

\usepackage{graphicx}
\usepackage{amsmath}
\usepackage{amssymb}
\usepackage{booktabs}
\usepackage[ruled,vlined]{algorithm2e}
\usepackage{graphicx}
\usepackage{booktabs}

\usepackage[accsupp]{axessibility}  
\usepackage{subcaption}
\usepackage[table]{xcolor}
\usepackage{svg}
\usepackage{booktabs}
\usepackage{comment}
\usepackage{tikz}
\usetikzlibrary{spy}
\usepackage{booktabs}
\usepackage{fonttable}
\usepackage{adjustbox}
\newcommand{\stkout}[1]{\ifmmode\text{\sout{\ensuremath{#1}}}\else\sout{#1}\fi}
\usepackage{amssymb}
\usepackage{pifont}
\newcommand{\xmark}{\ding{55}}%
\usepackage{tikz,pgfplots}
\usepackage{subcaption} 
\usepackage{multicol}
\usepackage{siunitx}
\usepackage{multirow}
\usepackage{soul}
\usepackage[normalem]{ulem}

\usepackage{times}
\usepackage{epsfig}
\usepackage{graphicx}
\usepackage{amsmath}
\usepackage{amssymb}
\usepackage{cancel}

\usepackage[pagebackref,breaklinks,colorlinks]{hyperref}

\usepackage[capitalize]{cleveref}
\crefname{section}{Sec.}{Secs.}
\Crefname{section}{Section}{Sections}
\Crefname{table}{Table}{Tables}
\crefname{table}{Tab.}{Tabs.}

\def\wacvPaperID{45} 
\def\confName{WACV}
\def\confYear{2025}

\begin{document}

\title{RETHINED: A New Benchmark and Baseline for Real-Time High-Resolution Image Inpainting On Edge Devices}

\author{Marcelo Sanchez\\
Crisalix\\
{\tt\small marcelo.sanchez@crisalix.com}
\and
Gil Triginer\\
Crisalix\\
{\tt\small gil.triginer@gmail.com}
\and
Ignacio Sarasua\\
NVIDIA\\
{\tt\small isarasua@nvidia.com}
\and
Lara Raad\\
IIE, FIng, UdelaR\\
{\tt\small lraad@fing.edu.uy}
\and
Coloma Ballester\\
UPF\\
{\tt\small coloma.ballester@upf.edu}
}

 \twocolumn[{
 \maketitle
 \renewcommand\twocolumn[1][]{#1}%
 \begin{center}
     \captionsetup{type=figure}
    \includegraphics[width=\textwidth]{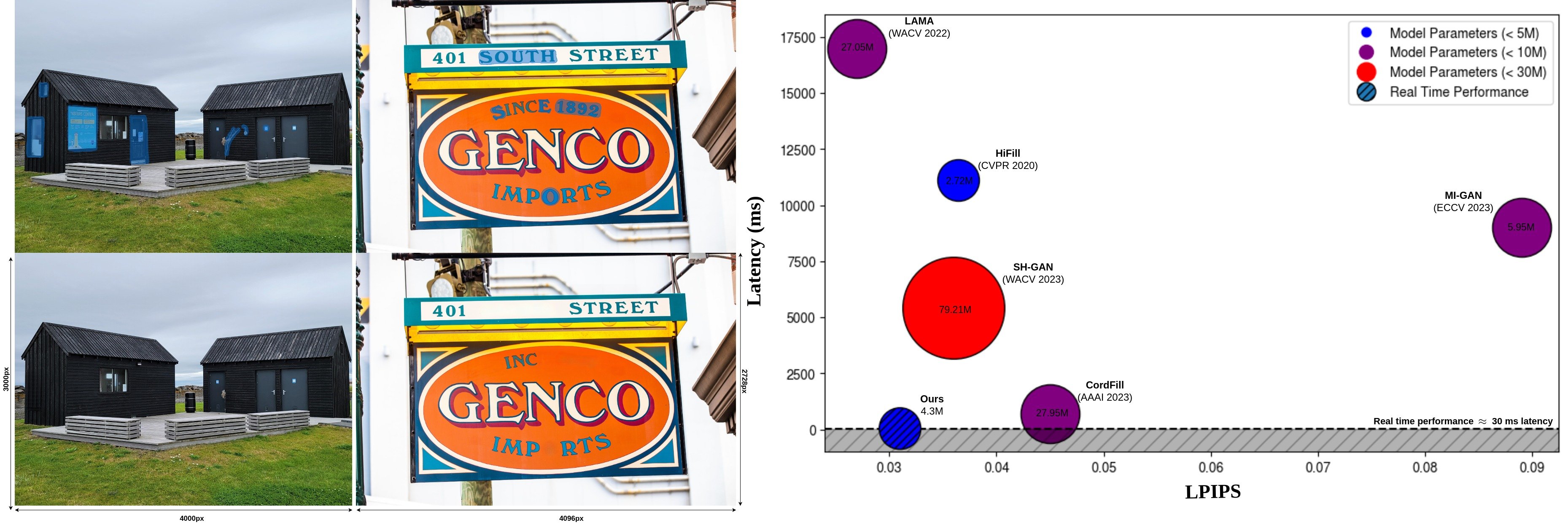}
    \vspace{-5mm}
     \captionof{figure}{Left: Inpainting result on ultra high-resolution images (best viewed by zoom-in on screen). Right: Comparison of LPIPS performance and Latency among different state-of-the-art methods.}\label{fig:teaser}
 \end{center}
 }]



\begin{abstract}
Existing image inpainting methods have shown impressive completion results for low-resolution images. However, most of these algorithms fail at high resolutions and require powerful hardware, limiting their deployment on edge devices. Motivated by this, we propose the first baseline for REal-Time High-resolution image INpainting on Edge Devices (RETHINED) that is able to inpaint at ultra-high-resolution and can run in real-time ($\leq$ 30ms) in a wide variety of mobile devices. A simple, yet effective novel method formed by a lightweight Convolutional Neural Network (CNN) to recover structure, followed by a resolution-agnostic patch replacement mechanism to provide detailed texture. Specially our pipeline leverages the structural capacity of CNN and the high-level detail of patch-based methods, which is a key component for high-resolution image inpainting. To demonstrate the real application of our method, we conduct an extensive analysis on various mobile-friendly devices and demonstrate similar inpainting performance while being $\mathrm{100 \times faster}$ than existing state-of-the-art 
methods. Furthemore,
we realease DF8K-Inpainting, the first free-form mask UHD inpainting dataset. Code and dataset \href{https://crisalixsa.github.io/rethined/}{here}. 
\end{abstract}


\section{Introduction}
\label{sec:intro}


\begin{figure*}[!ht]
    \vspace{-3mm}
    \centering
    \includegraphics[width=0.98\textwidth,scale=2]{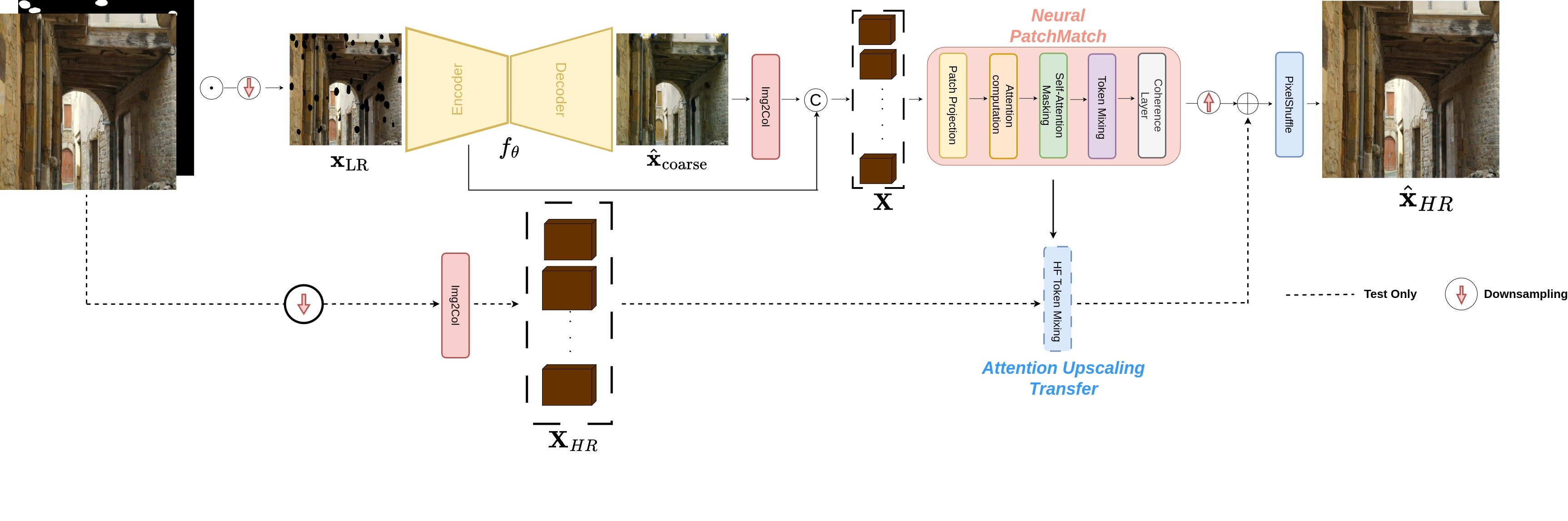}
    \vspace*{-6mm}
    \caption{\textbf{Proposed Inpainting Pipeline.} Given a HR image $\mathbf{y}$ and a binary mask $\mathbf{m}$ with corrupted pixels as inputs (left), our model first  downsamples $\mathbf{x} = \mathbf{y} \odot \mathbf{m}$ to a lower resolution $\mathbf{x}_{LR}$, and  forwards it to the coarse model $f_{\theta}$ obtaining   $\hat{\mathbf{x}}_{\text{coarse}}$. It is then refined by the NeuralPatchMatch module obtaining $\hat{\mathbf{x}}_{\text{LR}}$ and the attention map $\mathbf{A}$. From $\mathbf{A}$ and $\mathbf{x}$, our Attention Upscaling  module yields $\hat{\mathbf{x}}_{\text{HR}}$.}
    \label{fig:pipeline}
    \vspace{-3mm}
\end{figure*}


Image inpainting addresses the challenge of restoring or filling-in missing or damaged parts of an image, effectively ``completing'' the regions with plausible content. 
This process not only holds tremendous potential for photo restoration and editing but also finds applications in various domains, including art restoration, medical imaging and  video editing for media entertainment industry.


State-of-the-art methods have pushed the boundaries of image inpainting with deep convolutional techniques \cite{pathak2016context,yu2018generative,yu2019free,yi2020contextual,zeng2020high}, especially in the low-resolution (LR) regime (below 512 pixels in the longest axis). These models, in a fixed resolution, are able to generate meaningful content with complex structures and textures. Nevertheless, performance degrades as resolution increases \cite{kulshreshtha2022feature}, failing to inpaint with coherent semantics and not synthesizing high-detail textures, which appear naturally on high-resolution (HR) images. This is mainly due to the fact that their basic building block are Convolutional Neural Networks (CNN), which require very large receptive fields \cite{luo2016understanding} in order to simultaneously understand the high-level semantic structure, as well as the finely detailed textures, involved in HR inpainting. In plain CNNs a sufficiently large receptive field can only be achieved by stacking many convolutions, which is inefficient \cite{luo2016understanding}.
In order to avoid large networks, several solutions to this problem have been proposed. Suvorov et al. \cite{suvorov2022resolution} proposed large mask inpainting (LaMa) using Fast Fourier Convolutions (FFC) in the generator, which allows for an effective image-wide receptive field. However, with high-resolution images LaMa fails to inpaint with coherent content, introducing texture artifacts \cite{kulshreshtha2022feature, Zhang_2022_guided_pm}.



In contrast, classical non-local patch-based methods such as \cite{efros1999texture,demanet2003image,criminisi2004region,barnes2009patchmatch} are naturally able to inpaint high-detail textures by exploiting the available regularity and pattern redundancy with an explicit model governing the inpainting workflow \cite{aujol2010exemplar,arias2011variational,newson2014video,fedorov2016affine}. Nevertheless, they suffer of long computational time (despite the relative efficiency of PatchMatch \cite{barnes2009patchmatch}) and of the inability to incorporate  high-level semantic information. 


To marry the best of both worlds, we introduce a novel lightweight inpainting pipeline, based on small CNNs (local) and patch-based (global) methods for ultra high-resolution image completion. It allows to learn the underlying semantic structure while borrowing high-detail textures features from uncorrupted regions.

Our method leverages three important design principles: i) Use of CNNs to learn a coarse representation on LR images containing high-level semantic structure, ii) Patch-level matching to inpaint with high-detail textures from the known image regions, iii) Patch-agregation to allow consistency and robustness inpainting at 
super high-resolutions.

Our method, compared with other  state-of-the-art mobile inpainting methods, obtains similar reconstruction accuracy at LR images and clearly outperforms other state-of-the-art methods at HR while reducing order-of-magnitude latency inference cost (we refer to Table \ref{table:quantitative} and Fig.~\ref{fig:teaser}). Furthermore, our method has the ability to be consistent at different super high-resolutions while being trained on LR images (Fig. \ref{fig:importance_of_attention_upscaling}), reducing the cost of crafting a HR image dataset.

In order to demonstrate the real application of our methods, we perform an extensive analysis (Table \ref{table:inferencetimemultipledevices}) of inference speed on multiple edge devices from different chip manufacturers. Table \ref{table:inferencetimemultipledevices} demonstrates that our model is $100 \times$ faster than MI-GAN\cite{Sargsyan_2023_ICCV}, $1.7 \times$ faster than Coordfill \cite{liu2023coordfill} while obtaining similar reconstruction metrics on LR and outperforming in HR (Table \ref{table:quantitative}). Furthermore, our model has $8 \times$ less parameters, which improves drastically loading times on edge devices.


Moreover, we construct and release the first dataset of free-form masks for super high-resolution image inpainting.  Previous work \cite{Zhang_2022_guided_pm} only released a small test set, making it difficult to train entirely end-to-end high-resolution inpainting models without the need of additional private data.
Summarizing, our contributions are as follows:\\ 
$\bullet$ A novel lightweight pipeline for real-time image inpainting that enables ultra-high-resolution image completion on edge devices with limited memory. The model can be efficiently deployed in a wide variety of edge devices.\\
$\bullet$ A novel attention upscaling module that generalizes to several ultra-high-resolutions while being trained on LR.\\
$\bullet$ DF8K-Inpainting, the largest free-form mask inpainting dataset for evaluating super high-resolution methods.

\begin{figure*}
\begin{tabular}{cccccc}
\centering
\hspace*{-0.2in}
\begin{tikzpicture}[spy using outlines={thick,red,rectangle,magnification=5,size=2.cm,connect spies}]
    \centering
    \node[rectangle,draw,inner sep=0.01pt] (image) at (0,0){\includegraphics[height=0.25\linewidth,width=0.25\linewidth]{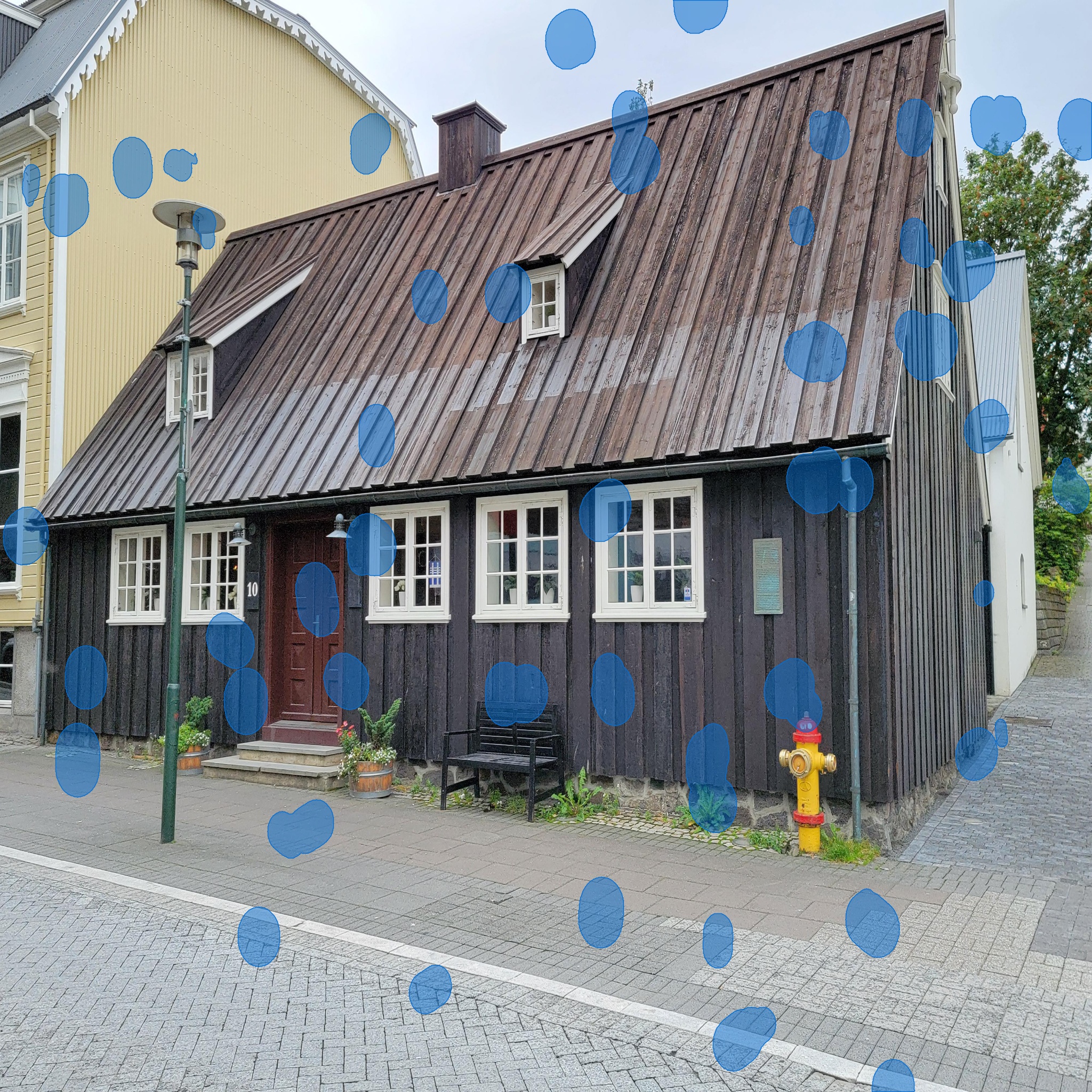}};
    \node [anchor=south, rotate=90, text opacity=1] at (image.west) {\normalsize $2048 \times 2048$};
    \spy[spy connection path={
}]on (-0.45,1.0) in node at (1.2,-1.2);
\end{tikzpicture}&

\hspace*{-0.2in}
\begin{tikzpicture}[spy using outlines={thick,red,rectangle,magnification=5,size=2.cm,connect spies}]
    \centering
    \node[rectangle,draw,inner sep=0.01pt] (image) at (0,0){\includegraphics[height=0.25\linewidth,width=0.25\linewidth]{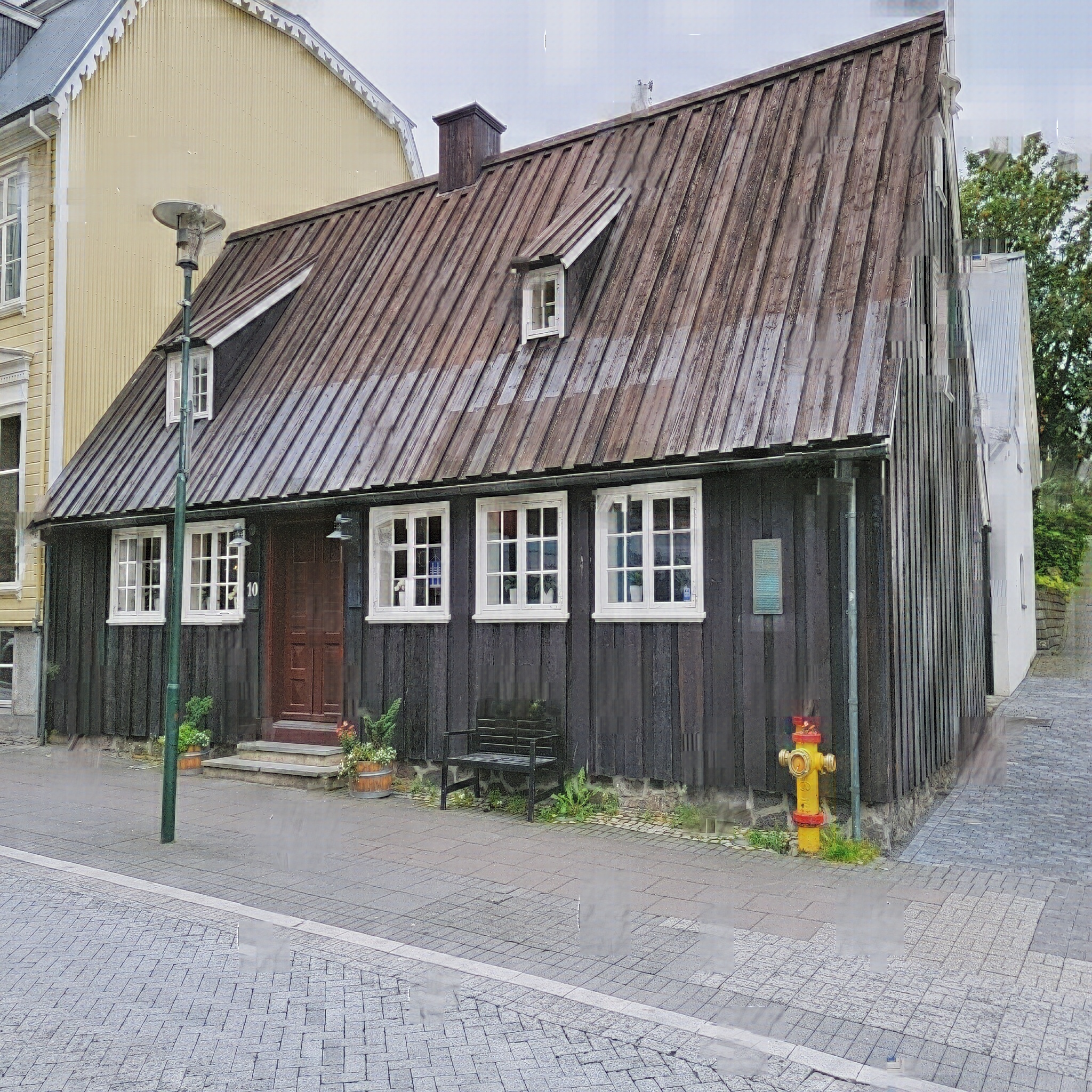}};
    \spy[spy connection path={
}]on (-0.45,1.0) in node at (1.2,-1.2);
\end{tikzpicture}&

\hspace*{-0.2in}
\begin{tikzpicture}[spy using outlines={thick,red,rectangle,magnification=5, size=2.cm,connect spies}]
    \centering
    \node[rectangle,draw,inner sep=0.01pt] (image) at (0,0){\includegraphics[height=0.25\linewidth,width=0.25\linewidth]{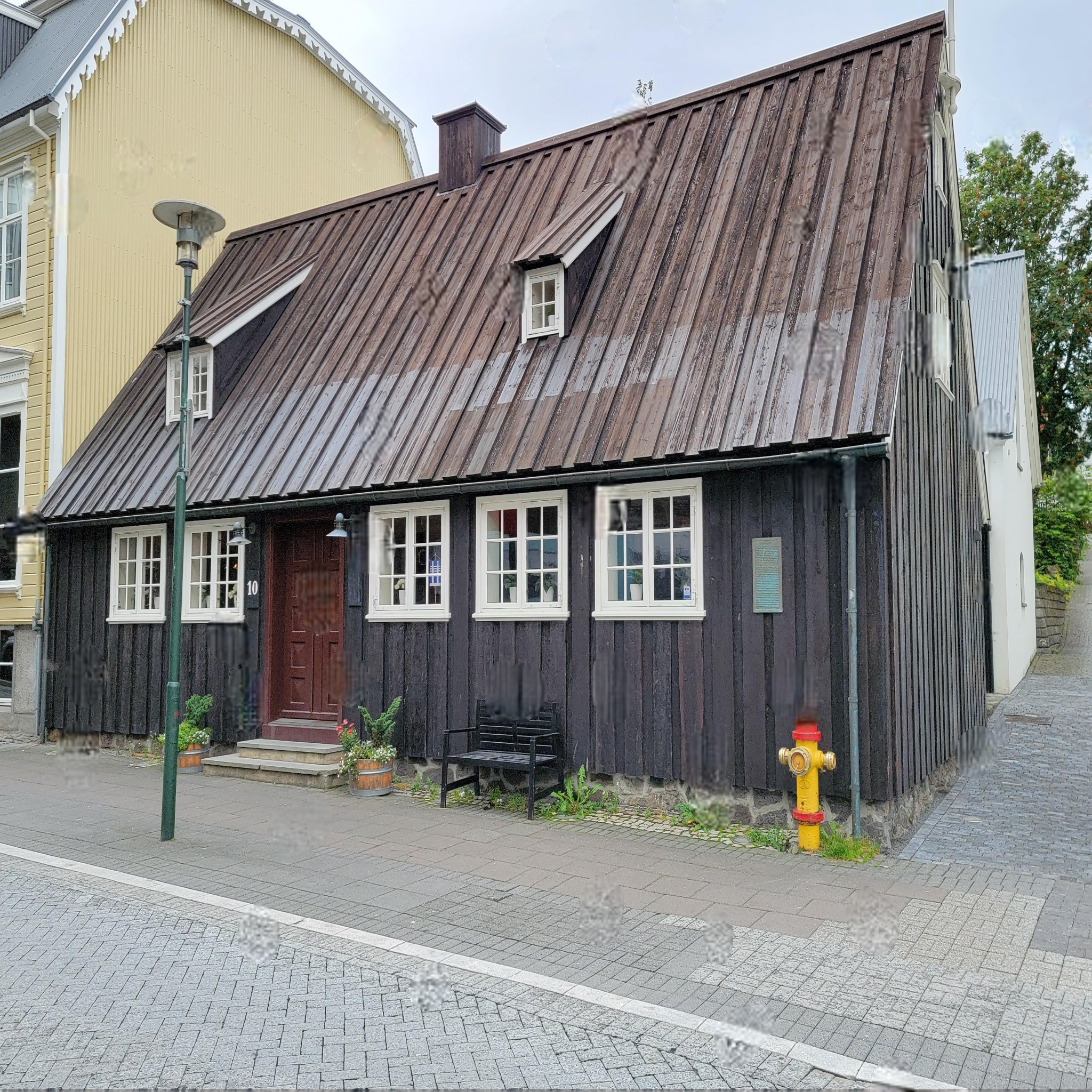}};
    \spy[spy connection path={
}]on (-0.45,1.0) in node at (1.2,-1.2);
\end{tikzpicture}&

\hspace*{-0.2in}
\begin{tikzpicture}[spy using outlines={thick,red,rectangle,magnification=5, size=2.cm,connect spies}]
    \centering
    \node[rectangle,draw,inner sep=0.01pt] (image) at (0,0){\includegraphics[height=0.25\linewidth,width=0.25\linewidth]{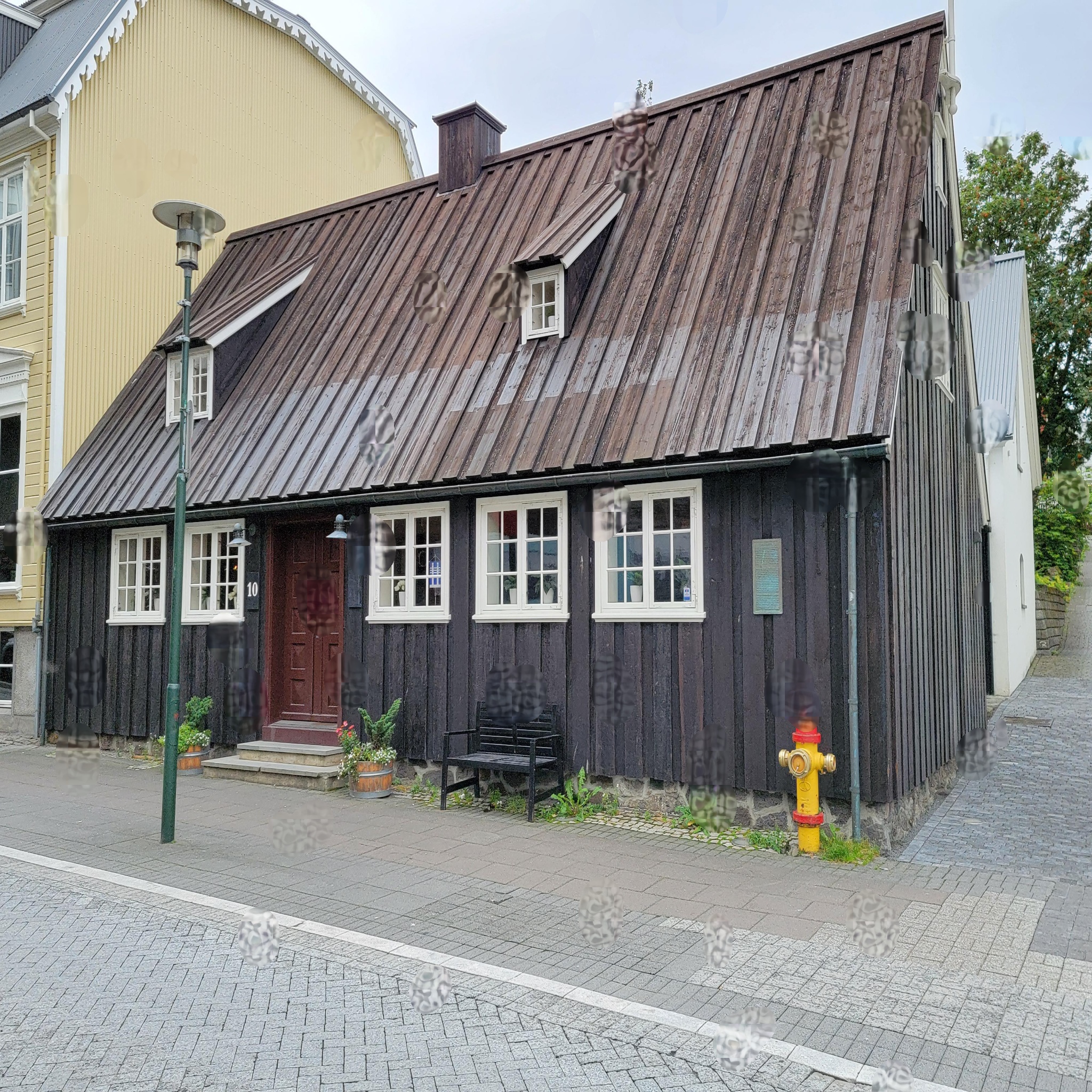}};
    \spy[spy connection path={
}]on (-0.45,1.0) in node at (1.2,-1.2);
\end{tikzpicture}& \\

\hspace*{-0.2in}
\begin{tikzpicture}[spy using outlines={thick,red,rectangle,magnification=5, size=2cm,connect spies}]
    \centering
    \node[rectangle,draw,inner sep=0.1pt] (image) at (0,0){\includegraphics[height=0.25\linewidth,width=0.25\linewidth]{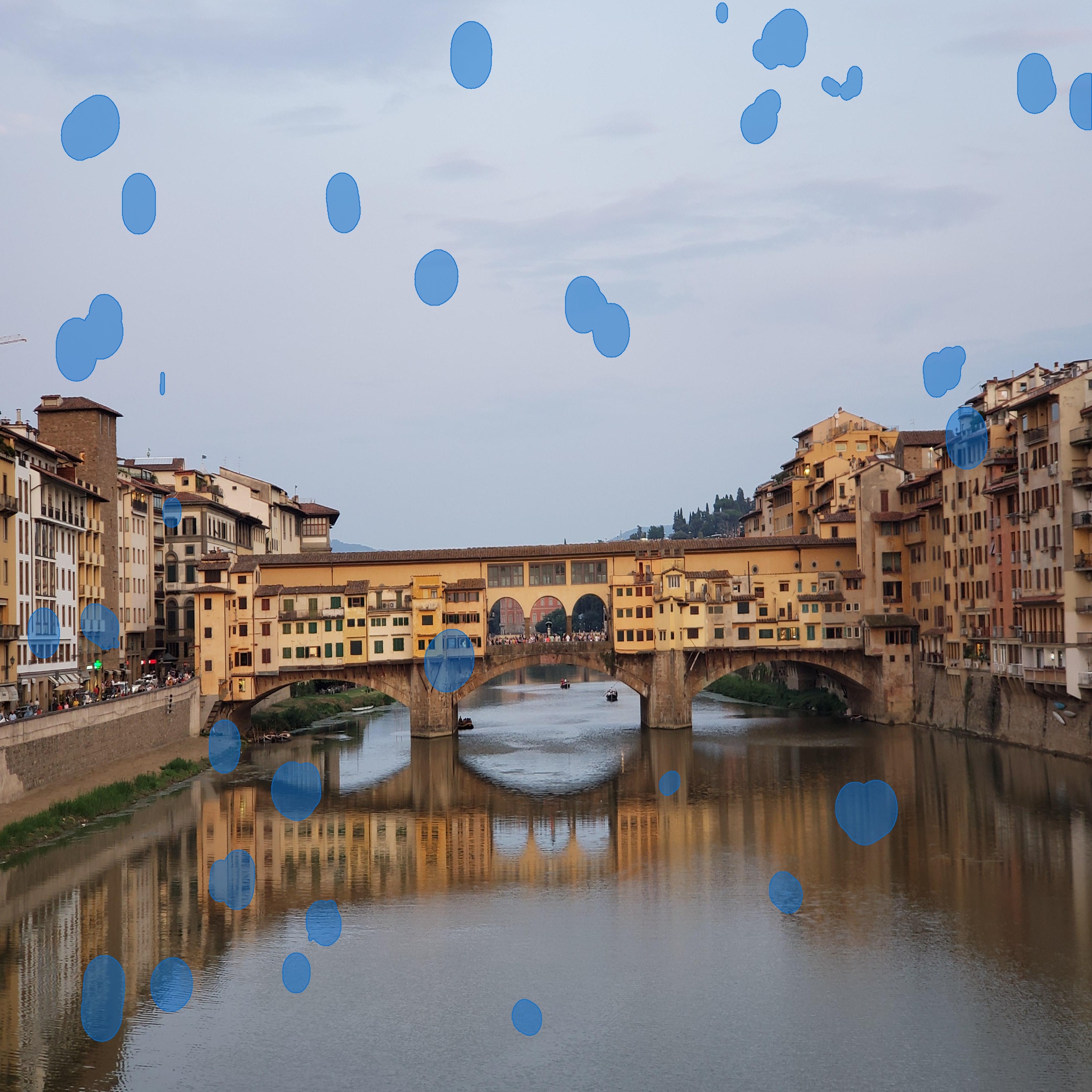}};
    \node [anchor=south, rotate=90, text opacity=1] at (image.west) {\normalsize $1024 \times 1024$};
    \spy[spy connection path={
}]on (-1.2,-1.3) in node at (1.2,-1.2);
\end{tikzpicture}&

\hspace*{-0.2in}
\begin{tikzpicture}[spy using outlines={thick,red,rectangle,magnification=5, size=2cm,connect spies}]
    \centering
    \node[rectangle,draw,inner sep=0.1pt] (image) at (0,0){\includegraphics[height=0.25\linewidth,width=0.25\linewidth]{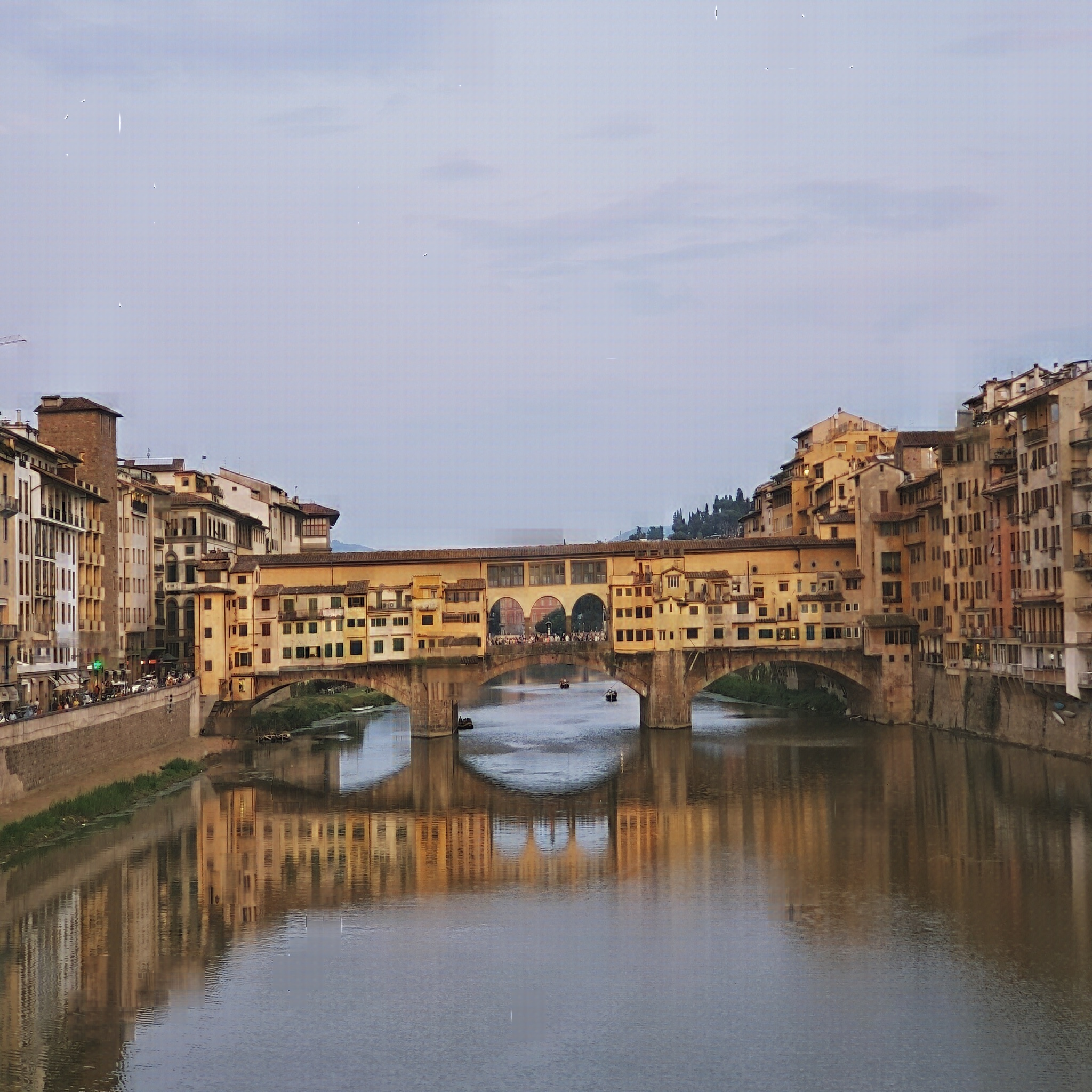}};
    \spy[spy connection path={
}]on (-1.2,-1.3) in node at (1.2,-1.2);
\end{tikzpicture}&

\hspace*{-0.2in}
\begin{tikzpicture}[spy using outlines={thick,red,rectangle,magnification=5, size=2cm,connect spies}]
    \centering
    \node[rectangle,draw,inner sep=0.1pt] (image) at (0,0){\includegraphics[height=0.25\linewidth,width=0.25\linewidth]{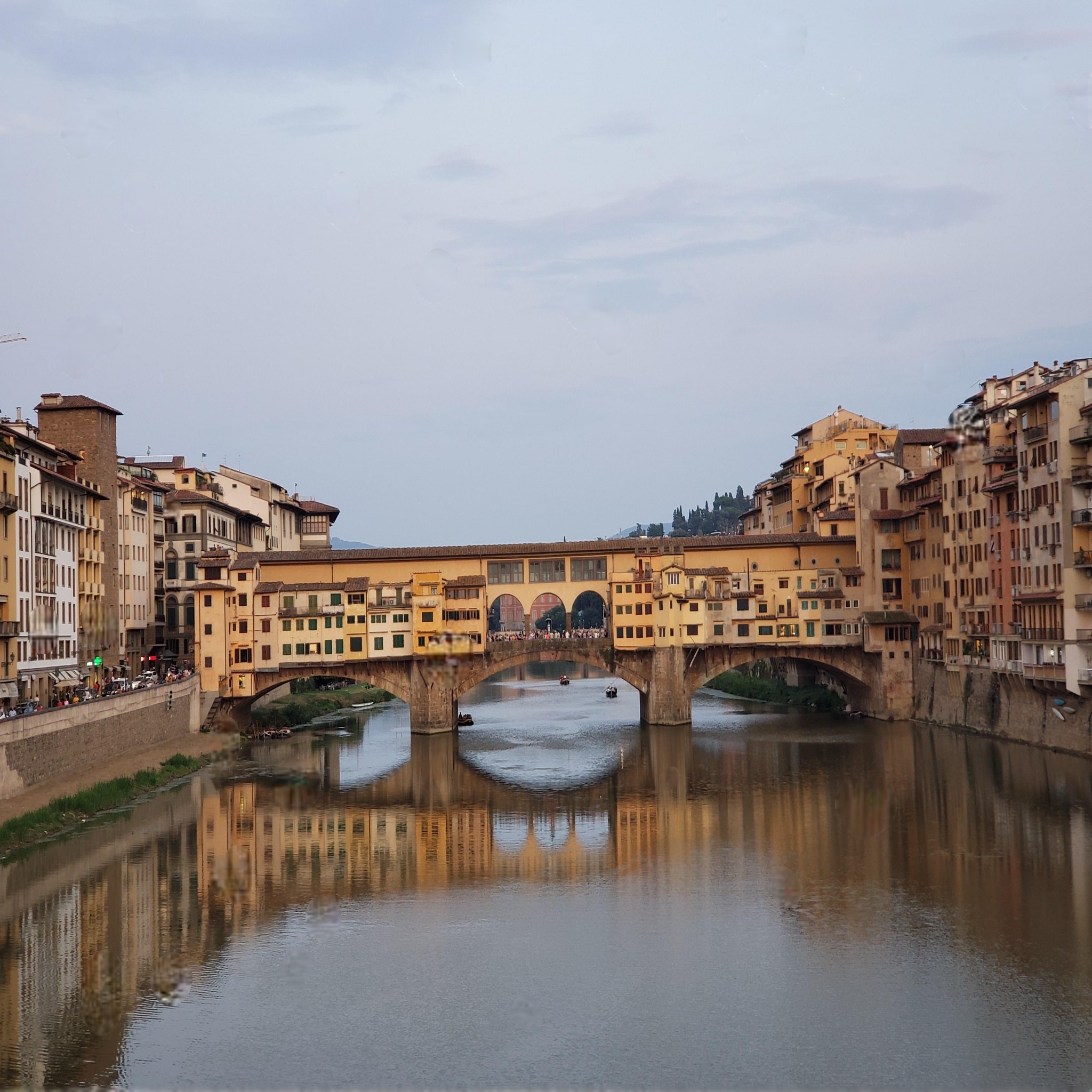}};
    \spy[spy connection path={
}]on (-1.2,-1.3) in node at (1.2,-1.2);
\end{tikzpicture}&

\hspace*{-0.2in}
\begin{tikzpicture}[spy using outlines={thick,red,rectangle,magnification=5, size=2cm,connect spies}]
    \centering
    \node[rectangle,draw,inner sep=0.1pt] (image) at (0,0){\includegraphics[height=0.25\linewidth,width=0.25\linewidth]{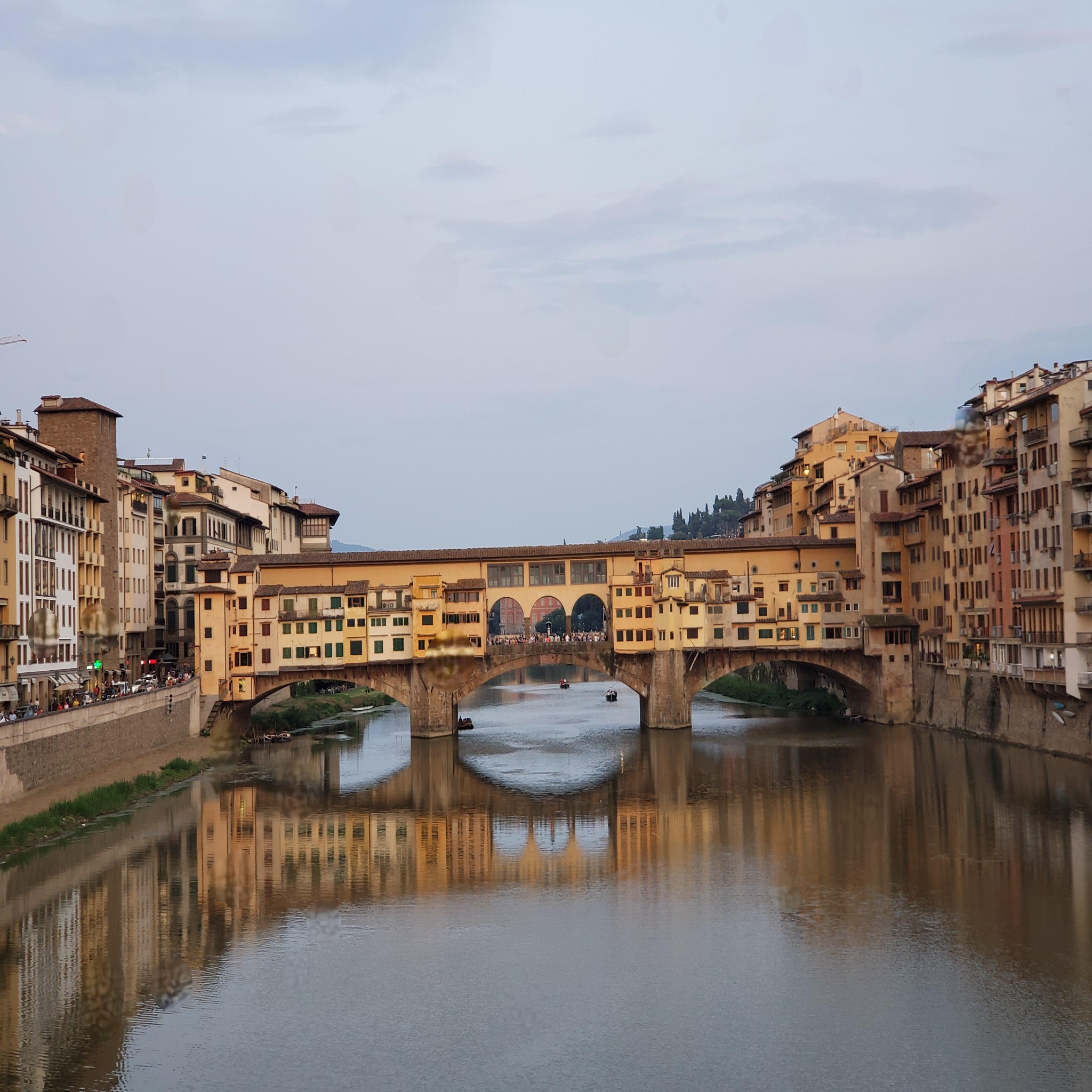}};
    \spy[spy connection path={
}]on (-1.2,-1.3) in node at (1.2,-1.2);
\end{tikzpicture} & \\

\hspace*{-0.2in}
\begin{tikzpicture}[spy using outlines={thick,red,rectangle,magnification=5, size=2cm,connect spies}]
    \centering
    \node[rectangle,draw,inner sep=0.1pt] (image) at (0,0){\includegraphics[height=0.25\linewidth,width=0.25\linewidth]{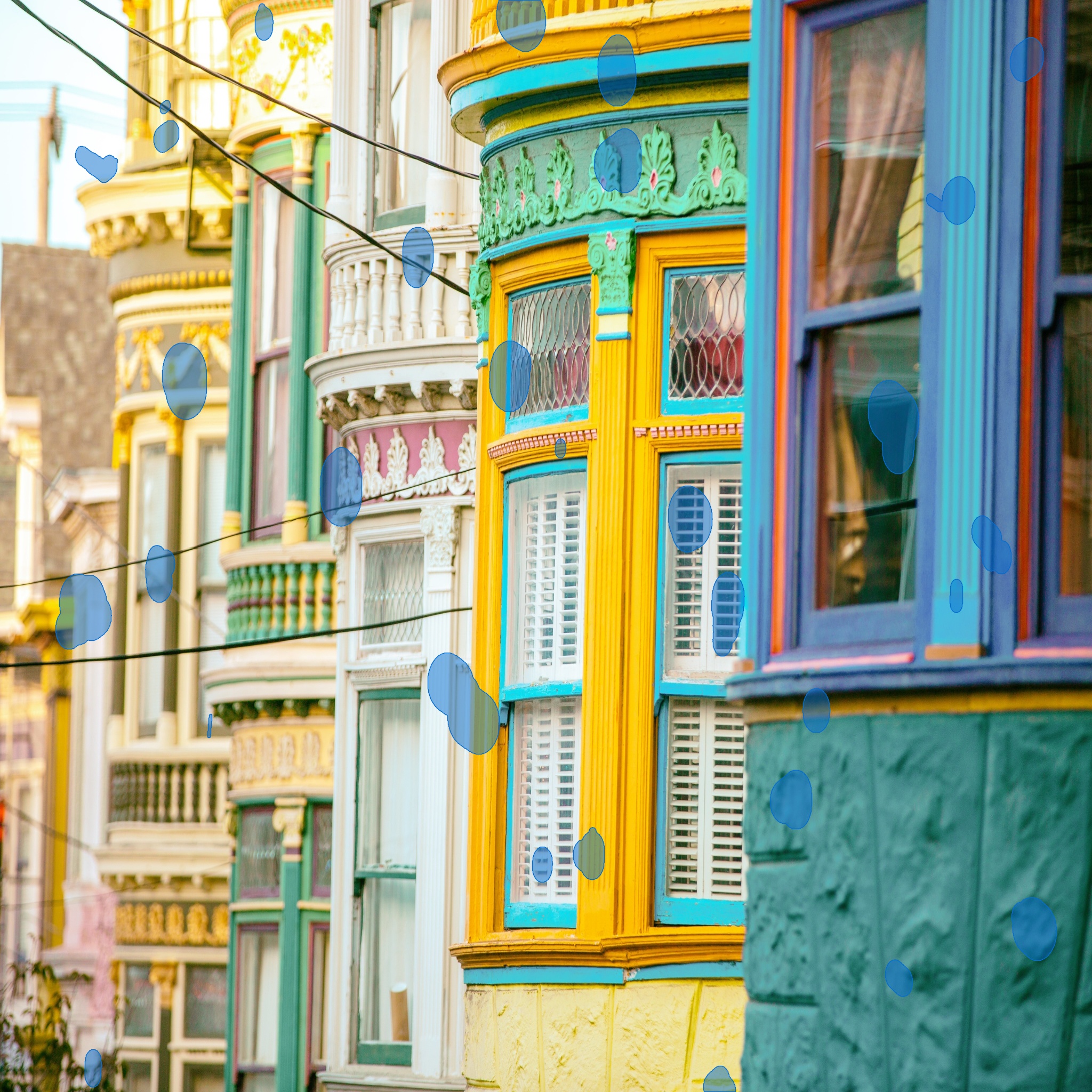}};
    \node [anchor=south, rotate=90, text opacity=1] at (image.west) {\normalsize $4096 \times 4096$};
    \spy[spy connection path={
}]on (0.6,-0.3) in node at (1.2,-1.2);
\end{tikzpicture} &

\hspace*{-0.2in}
\begin{tikzpicture}[spy using outlines={thick,red,rectangle,magnification=5, size=2cm,connect spies}]
    \centering
    \node[rectangle,draw,inner sep=0.1pt] (image) at (0,0){\includegraphics[height=0.25\linewidth,width=0.25\linewidth]{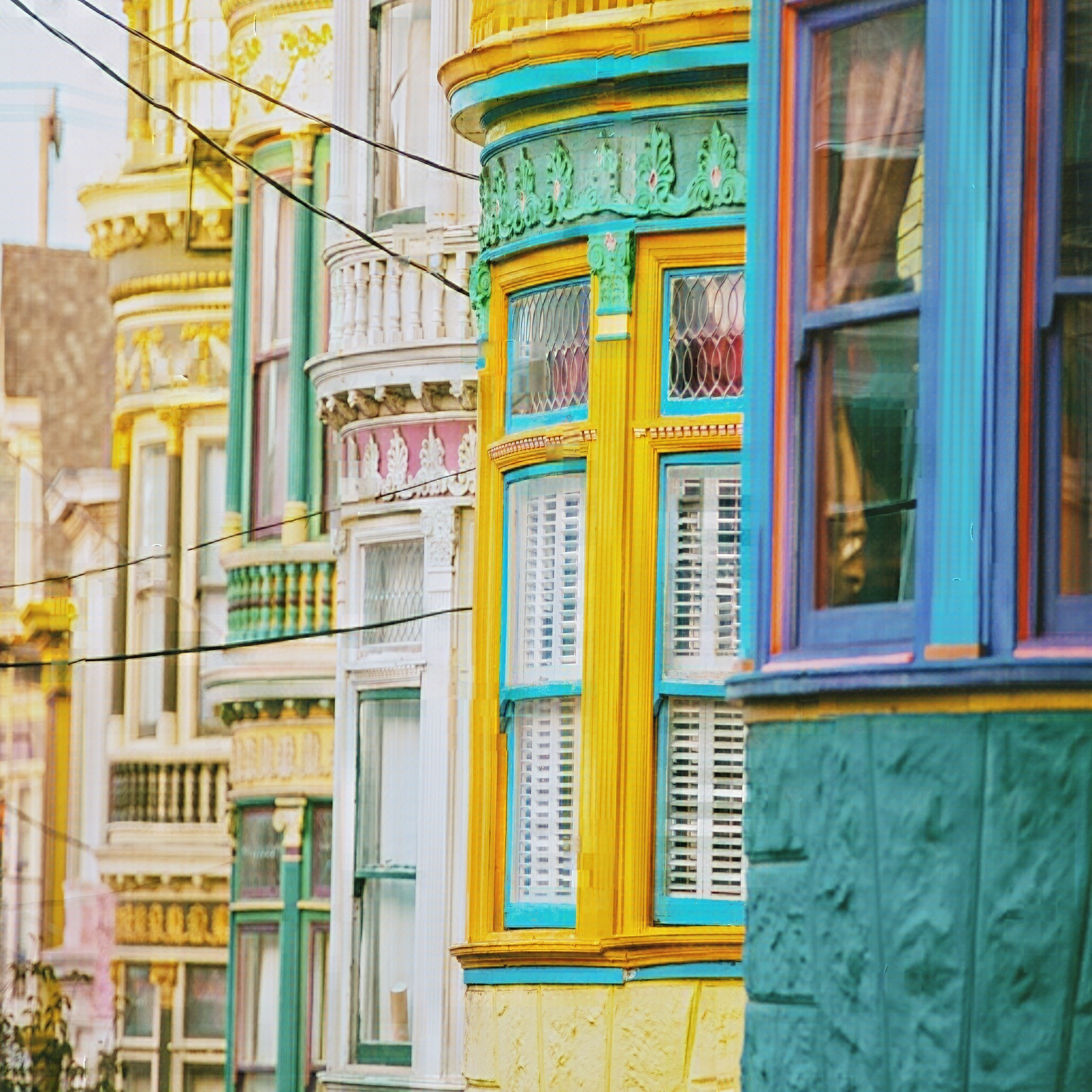}};
    \spy[spy connection path={
}]on (0.6,-0.3) in node at (1.2,-1.2);
\end{tikzpicture} &

\hspace*{-0.2in}
\begin{tikzpicture}[spy using outlines={thick,red,rectangle,magnification=5, size=2cm,connect spies}]
    \centering
    \node[rectangle,draw,inner sep=0.1pt] (image) at (0,0){\includegraphics[height=0.25\linewidth,width=0.25\linewidth]{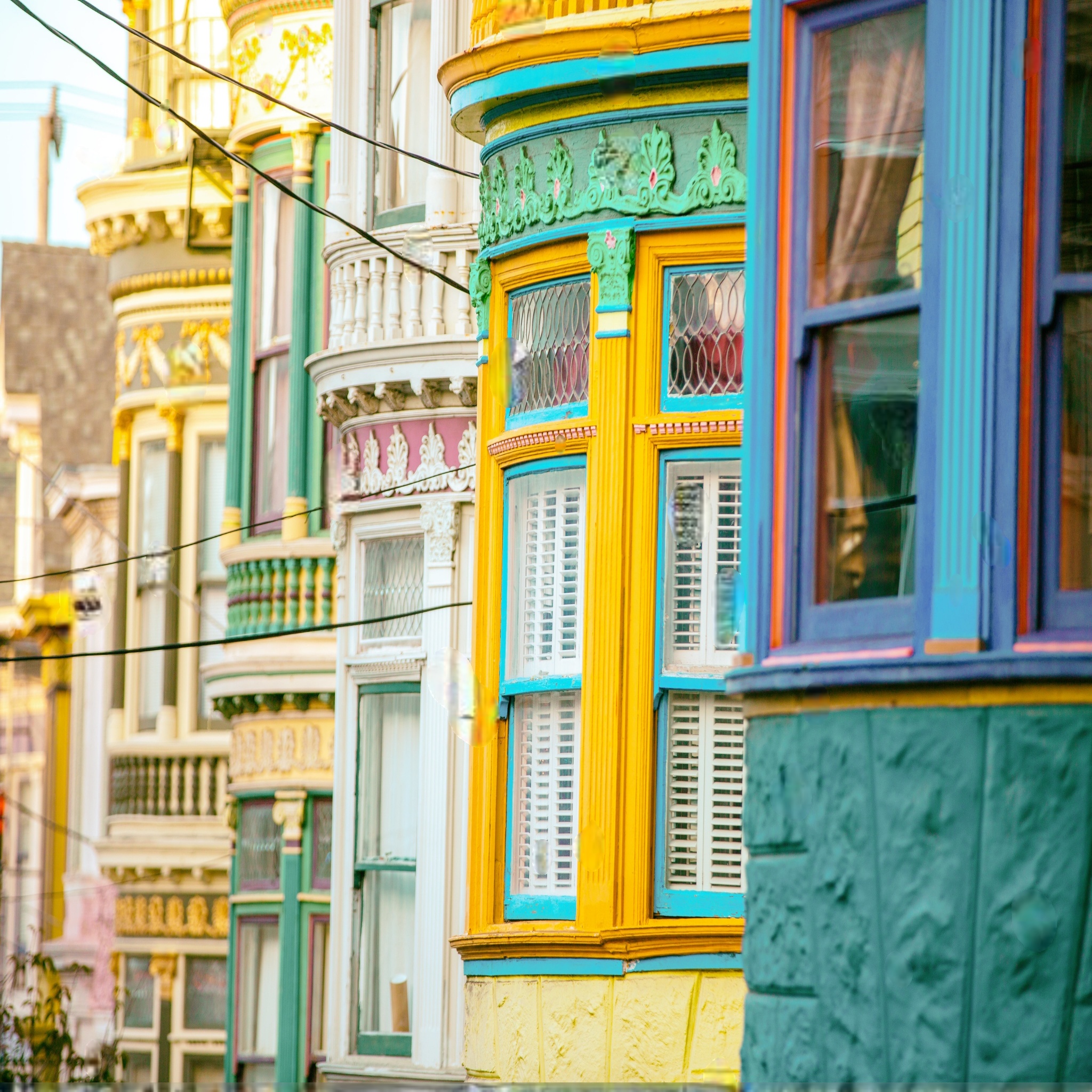}};
    \spy[spy connection path={
}]on (0.6,-0.3) in node at (1.2,-1.2);
\end{tikzpicture} &

\hspace*{-0.2in}
\begin{tikzpicture}[spy using outlines={thick,red,rectangle,magnification=5, size=2cm,connect spies}]
    \centering
    \node[rectangle,draw,inner sep=0.1pt] (image) at (0,0){\includegraphics[height=0.25\linewidth,width=0.25\linewidth]{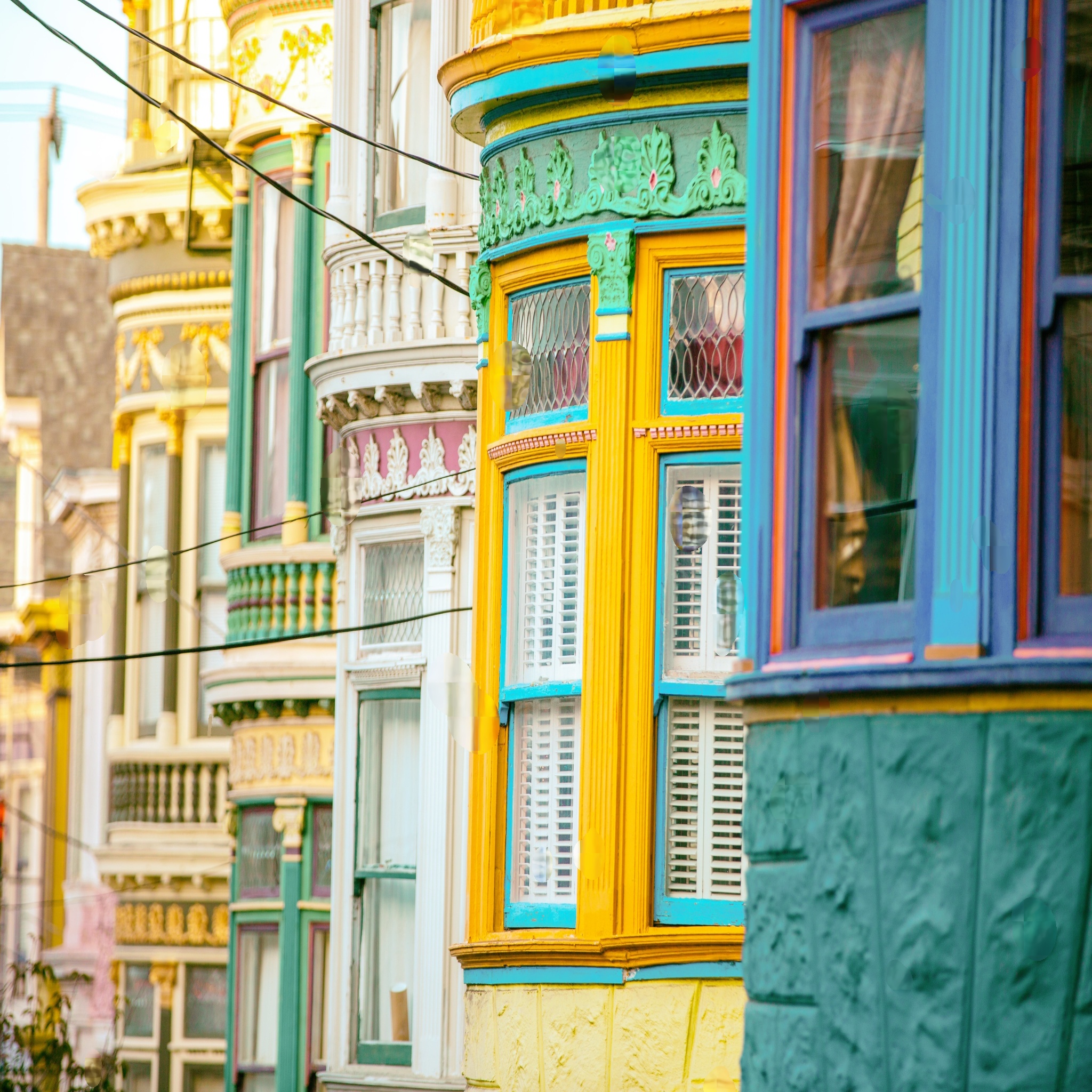}};
    \spy[spy connection path={
}]on (0.6,-0.3) in node at (1.2,-1.2);
\end{tikzpicture} & \\

{{Masked Image}} & {{Ours} \textbf{(34.33 ms)}} & {{MI-GAN~\cite{Sargsyan_2023_ICCV} \textbf{(3905.4 ms)}}} & {{CoorFill~\cite{liu2023coordfill} \textbf{(101.4 ms)}}}
\end{tabular}

\caption{Comparison of different inpainting methods able to work on mobile devices. Latency speed appears in parentheses and has been calculated at $2048 \times 2048$ resolution on Apple M2 Ipad Pro.}
\label{fig:results_inpainting_FFHQ}
\end{figure*}
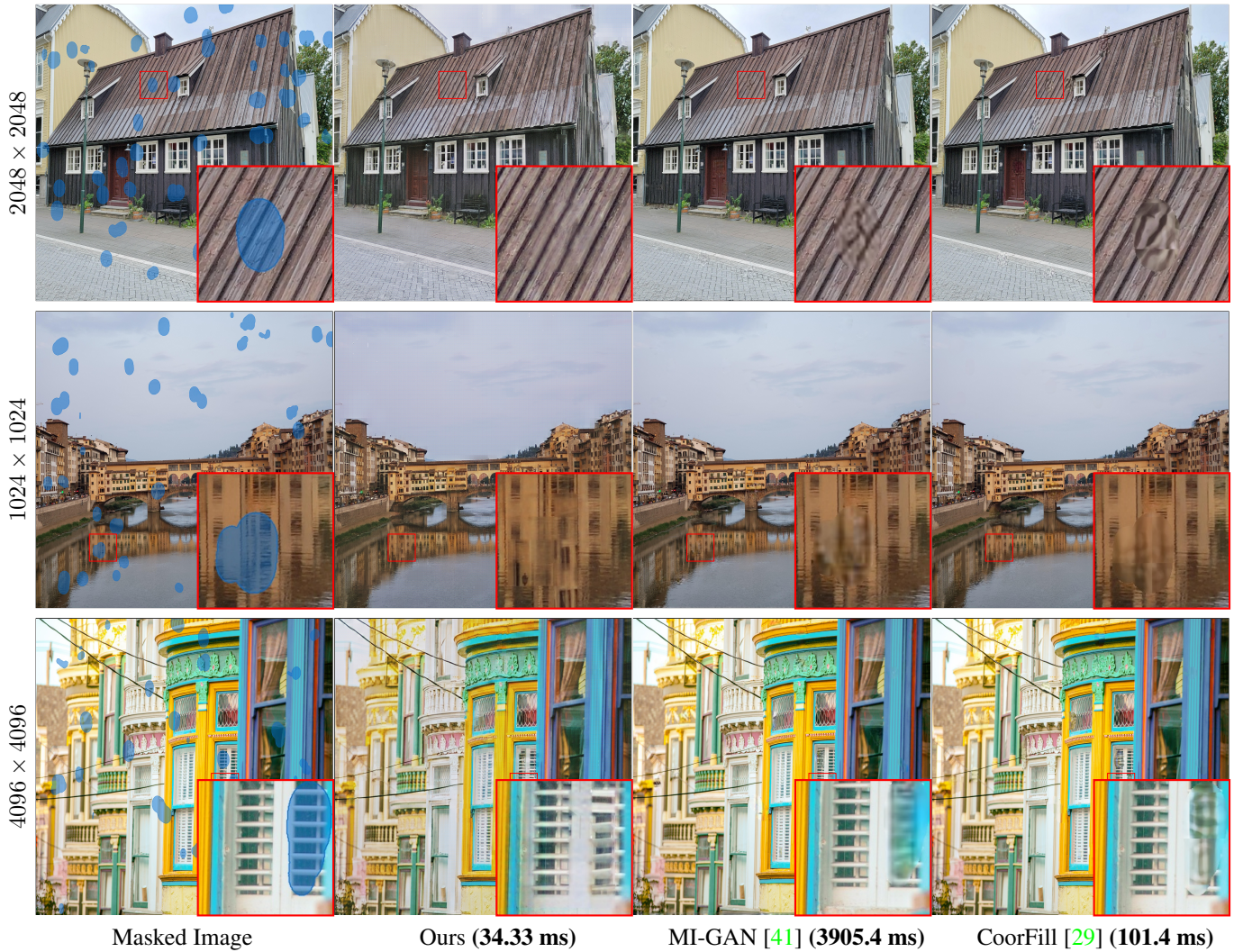
 \section{Related Work}
\label{sec:relatedwork}

\begin{figure*}[!ht]
    \centering
    \includegraphics[width=0.98\textwidth,scale=2]{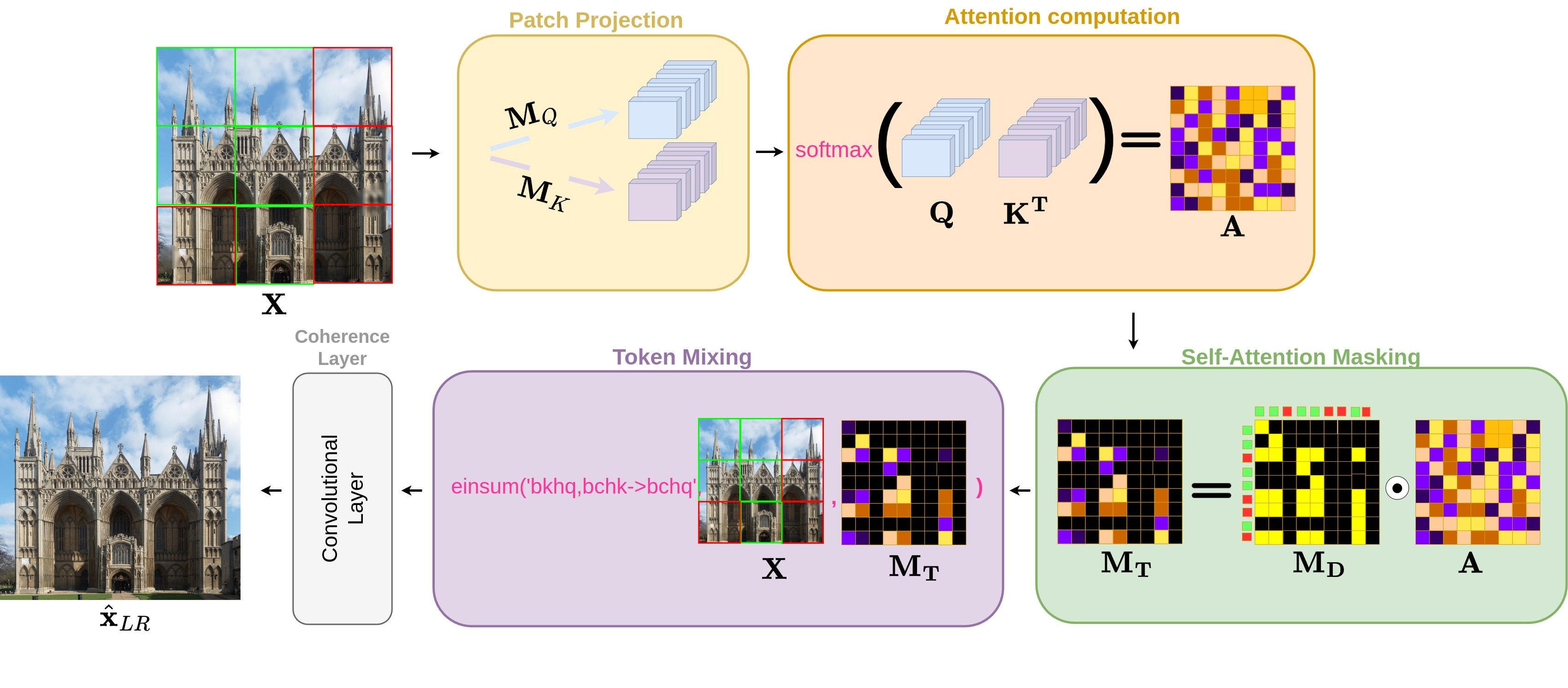}
    \vspace*{-7mm}
    \caption{\textbf{Proposed NeuralPatchMatch Inpainting Module.} (Corrupted patches are displayed as red \fcolorbox{red}{white}{\rule{0pt}{5pt}\rule{5pt}{0pt}} while uncorrupted ones as green \fcolorbox{green}{white}{\rule{0pt}{5pt}\rule{5pt}{0pt}}.) First, we project patch embedding to embedding space of dimension $d_{k}$ (Sect.~\ref{sec:neural_patch_match}). Then token similarity is computed in a self-attention manner, obtaining attention map $\mathbf{A}$ (where lighter colors \fcolorbox{black}{yellow}{\rule{0pt}{5pt}\rule{5pt}{0pt}} correspond to a large softmax value while darker colors \fcolorbox{black}{black}{\rule{0pt}{5pt}\rule{5pt}{0pt}} correspond to a low value). The self-attention masking allows to inpaint only on corrupted regions, maintaining high-frequency details from uncorrupted zones. To obtain the final inpainted image, we mix the tokens via a weighted sum based on the attention map $\mathbf{A}$.}
    \label{fig:patchmatch}
    \vspace{-3mm}
\end{figure*}

\subsection{High-resolution Image Inpainting}

Most state-of-the-art inpainting  methods \cite{lu2022glama,dong2022incremental,li2022mat,cao2023zits++} are evaluated on low-resolution datasets such as Places2 \cite{zhou2017places} and CelebHQ \cite{karras2019style} which drastically differ with current camera sensors. Current mobile devices have sensors with lenses capable of capturing world above 2 MegaPixels (MP), while current image inpainting datasets are still evaluating under 1MP resolution. This difference is an underlying problem in order to use existing methods in real-world applications. Few works exist that evaluate specifically on high-resolution images. Old methods such as \cite{yang2017high} define high-resolutions above 512px, which is not today's idea of high-resolution.  \cite{zeng2020high} proposed a method able to inpaint up to 8K resolution. To do so, they learn the image structure at a lower resolution and use guidance to upsample the predicted features. \cite{zeng2020high} introduced an iterative method to progressively fill the corrupted regions. While these methods are able to inpaint with coherent information while preserving high-frequency detail, they do not meet the latency and memory constraints for proper deployment on edge devices.

\subsection{Real Time Mobile Image Inpainting}

Mobile devices, specifically smartphones, have become the primarily tool for world capture. These gadgets provide low energy limited hardware but posses high-resolution camera sensors able to capture high quality photos similar to professional cameras \cite{delbracio2021mobile}.

Moreover, mobile devices have been adopted as content editing platforms. Mobile software such as Snapseed \cite{Snapseed} or Adobe Lightroom \cite{lightroom} are largely used nowadays. These platforms employ algorithms such as object-removal, super-resolution or denoising on videos and images up to 4K resolution. Besides, these platforms require real-time performance to provide a fluent response.

Yet, to the best of our knowledge, no work exists for real-time high-resolution image inpainting on edge devices. As matter of fact, only few works exist in the inpainting field that consider inference latency as a constraint.

Liu et al \cite{liu2023coordfill} proposed a reduced LaMa inpainting network \cite{suvorov2022resolution} with a coordinate-based multi-layer perceptron. This approach helped to be robust against high resolution.  However, the latency performance is unacceptable and real-time can only be achieved with a modern desktop GPU. 

To the best of our knowledge, MI-GAN \cite{Sargsyan_2023_ICCV} is the only work focused on inpainting for mobile devices deployment. However, its inference results are two orders of magnitude away from real-time inference, and furthermore, it is not robust against high-resolution images,  only obtaining coherent completion on low resolution.

\subsection{Mobile-Friendly Neural Networks}

Development for building efficient mobile-friendly Neural Networks (NN) has seen a lot of progress in recent years \cite{ma2018shufflenet,vasu2023mobileone,mehta2021mobilevit,sandler2018mobilenetv2,chen2022mobile,howard2019searching,mehta2021mobilevit}, setting a quorum of best practices for mobile-friendly neural networks.

Architecture designs such as kernel size in convolutions \cite{sandler2018mobilenetv2}, number of convolutional filters \cite{vasu2023mobileone}, network parallelism \cite{ma2018shufflenet}, activation functions \cite{vasu2023mobileone}, attention mechanism \cite{mehta2021mobilevit}, play an important role in performance on edge devices. It is vital to understand the underlying factors that improve efficiency in order to develop a general method that succeeds across many devices.

Table \ref{table:inferencetimemultipledevices} shows the comparison with different mobile-friendly inpainting networks that exist.

 \section{Method}
\subsection{Pipeline Overview}
Given a high-resolution RGB image $\mathbf{y} \in \mathbb{R}^{H_{\text{HR}}\times W_{\text{HR}}\times 3}$ (where $H_{\text{HR}}$ and $W_{\text{HR}}$ denote, respectively, the height and width of the high-resolution image in pixels) and a binary mask $\mathbf{m} \in \mathbb{R}^{H_{\text{HR}}\times W_{\text{HR}}}$ containing the corrupted pixels, our goal is to fill-in with plausible information the masked image $\mathbf{x} = \mathbf{y} \odot \mathbf{m}$.
To achieve this goal, we first downsample $\mathbf{x}$ to a lower resolution obtaining $\mathbf{x_{\text{LR}}}\in \mathbb{R}^{H\times W\times 3}$ (where $H<H_{\text{HR}}$ and $W<W_{\text{HR}}$) and forward it to the coarse model, obtaining the coarse inpainted image $\hat{\mathbf{x}}_{\text{coarse}}$ of size $H\times W$ (Sect. \ref{sec:coarseInp}). Then, we use the NeuralPatchMatch module (Sect. \ref{sec:neural_patch_match}) to refine $\hat{\mathbf{x}}_{\text{coarse}}$ by propagating known content from the input image $\mathbf{x_{\text{LR}}}$, obtaining $\hat{\mathbf{x}}_{\text{LR}}$ and the corresponding attention map $\mathbf{A}$.
Finally our Attention Upscaling  module (Sect. \ref{sec:attention_upscaling}) uses the learned attention map $\mathbf{A}$ and $\mathbf{x}$ to resemble high texture details found on the base image, finally obtaining a high-resolution image $\hat{\mathbf{x}}_{\text{HR}}$. The entire pipeline is displayed in Fig.~\ref{fig:pipeline}.
\subsection{Coarse Inpainting}\label{sec:coarseInp}

CNNs have shown exceptional results learning high-level semantic structure on low-resolution image inpainting \cite{Sargsyan_2023_ICCV,liu2023coordfill,karras2019style}. Based upon this, we employ a lightweight CNN $f_{\theta}(\cdot)$  that will produce a low resolution coarse inpainting. To do so, we first downsample the input image to $H\times W$ resolution and forward it through the network $f_{\theta}(\cdot)$ that fills the holes coarsely, obtaining an initial guess for the final inpainting $\hat{\mathbf{x}}_{\text{coarse}}=f_{\theta}(\mathbf{x}_{\text{LR}})\in \mathbb{R}^{H\times W\times 3}$. To ensure real-time performance, we have opted for a simple encoder-decoder network \cite{vasu2023mobileone, ronneberger2015u}. The network $f_{\theta}(\cdot)$ is formed by 5 blocks built upon $3 \times 3$ depthwise convolutions, 1x1 pointwise convolutions and batch normalization \cite{ioffe2015batch}. 
Specific details on the coarse model can be found in the supplementary material.
This architecture design allows all skip connections to be re-parameterized during inference. This not only avoids excessive network branching, but also enables low-latency performance.

\subsection{NeuralPatchMatch}
\label{sec:neural_patch_match}
Inspired by the self-attention mechanism \cite{vaswani2017attention} and the patch match strategy~\cite{barnes2009patchmatch}, we design a novel method that can pick known image features as reference to fill missing regions by incorporating a global receptive field and high texture quality restoration with as less computations as possible.  We enable long range dependencies by a neural matching procedure that is able to propagate information from the entire image as shown in Fig. \ref{fig:patchmatch}.

\noindent\textbf{Patch Embedding \& Projection.} 
Following a similar approach to \cite{dosovitskiy2020image} for patch embedding, we first split $\hat{\mathbf{x}}_{\text{coarse}}$ into $N$ non-overlapping square patches, $\mathbf{p}_{i}, i=1,\dots,N$. We do so via an optimized \texttt{img2col} operation (details on  the supplementary material) obtaining $\mathbf{I_{p}} = [\mathbf{p}_{1},...,\mathbf{p}_{N}]$, where $N = HW/P^2$ is the number of patches (sequence length), $P$ is the  side size of the defined patches and $\mathbf{p}_{i} \in \mathbb{R}^{3P^2}$ for $i=1,\dots,N$ . Similarly, we split the binary mask $\mathbf{m}$, obtaining  $\mathbf{M} =[m_1,...,m_N]$, where $m_i$ equals $1$ if any pixel of patch $\mathbf{p}_{i}$ is corrupted, $0$ otherwise. Furthermore, to impose structural information, which can not be easily extracted from plain RGB representation, we condition the attention on the intermediate features $F_{i}$ of our network $f_{\theta}(\cdot)$. Given the extracted intermediate features  $F_{i}$, at stage $i$, we concatenate them to $\mathbf{I_{p}}$, obtaining $\mathbf{X}$. The sequence of tokens $\bf{I_p}$ can be seen as a feature map of size $H/P\times W/P \times d_k$, were $d_{k}$ is the hidden dimension, to which we concatenate along the channel dimension the features of size $H/P\times W/P \times C$ from the coarse model $f_\theta$, thus obtaining $\bf{X}$ of size $H/P\times W/P \times (d_k + C)$.
 
\noindent\textbf{Attention Computation.} This sequence of input tokens $\mathbf{X}$ is forwarded trough equation \eqref{eq:attention} in a self-attention manner,
%
\begin{equation}
\text{Attention}(\mathbf{Q}, \mathbf{K}) = \text{softmax}\left(\frac{\mathbf{QK^T}}{\sqrt{d_k}}\right)
    \label{eq:attention}
\end{equation}
where $\mathbf{Q}, \mathbf{K}$ are the token projection over matrix $\mathbf{M}_{Q} \in \mathbb{R}^{N \times d_{k}}$ and $\mathbf{M}_{K} \in \mathbb{R}^{N\times d_{k}}$, respectively. We obtain an attention score matrix $\mathbf{A} \in \mathbb{R}^{N\times N}$

\noindent
\textbf{Self-Attention Masking.}
Even though image inpainting focus primarily on filling with coherent content the corrupted regions, it is also important to keep uncorrupted regions unchanged. In order to force the network to fill with new content only the masked regions and preserve original uncorrupted content, we use a binary mask $\mathbf{M_{D}}$ on the attention map $\mathbf{A}$. $\mathbf{M_{D}}$ maintains original uncorrupted patches and forces corrupted patches to attend uncorrupted patches. 
Finally masking the attention map $\mathbf{A}$ with  $\mathbf{M_D}$ results in the masked attention map $\mathbf{M_T} = \mathbf{A}  \cdot \mathbf{M_D}$, where $\mathbf{A}  \cdot \mathbf{M_D}$ denotes the element-wise product of the $N\times N$ matrices. Fig. \ref{fig:patchmatch} outlines this operation.

\noindent\textbf{Token Mixing.}
Once the similarity between tokens projection is encoded in $\mathbf{M_T}$, we communicate information among tokens to obtain the final inpainted image $\hat{\mathbf{x}}_{\text{LR}}$. To do so, we do a weighted contribution of the corresponding patches $\mathbf{q}_j$, where $\mathbf{q}_j$ can be either the LR patches or the HR patches with a weighting proportional to the attention map. That is, we obtain the reconstructed patches as:
\begin{equation}
\hat{\mathbf{p}}_{i}= \sum_j{M_T(i,j)\mathbf{q_{j}}}
 \label{eq:weighted_sum}
\end{equation}

\noindent\textbf{Coherence Layer.}
In order to avoid discontinuities between patch boundaries, we introduce a fast light-weight layer. This layer is based on simple yet effective convolutional filter that alleviates transitions between patches.

\noindent\textbf{Pixel Shuffle.}
To rearrange the sequence of reconstructed patches $\hat{\mathbf{p}}_{i}$ of dimension $(N,3 P^2)$ into an RGB image $\hat{\mathbf{x}}_{\text{LR}}$ of size $(H,W,3)$, we use the efficient low-memory footprint approach \cite{sze2017efficient}, PixelShuffle \cite{shi2016real}.

\begin{table*}[ht!]
\centering
\begin{adjustbox}{width=1\textwidth}
\begin{tabular}{lccccccccccccccccccccc}
\toprule
&  \multicolumn{9}{c}{1024x1024} & \multicolumn{5}{c}{2048x2048} & \multicolumn{5}{c}{4096x4096} \\
\cmidrule(lr{0.5em}){2-10} \cmidrule(lr{0.5em}){11-15} \cmidrule(lr{0.5em}){16-20}
& \multicolumn{4}{c}{CelebHQ} & \multicolumn{4}{c}{DIV8K} & & \multicolumn{4}{c}{DIV8K} & & \multicolumn{4}{c}{DIV8K}  \\
\cmidrule(lr{0.5em}){2-5} \cmidrule(lr{0.5em}){6-9} \cmidrule(lr{0.5em}){11-14} \cmidrule(lr{0.5em}){16-19}
& LPIPS $\downarrow$  & L1 $\downarrow$ & FID $\downarrow$ & SSIM $\uparrow$ & LPIPS $\downarrow$  & L1 $\downarrow$ & FID $\downarrow$ & SSIM $\uparrow$ & time (ms) $\downarrow$ & LPIPS $\downarrow$  & L1 $\downarrow$ & FID $\downarrow$ & SSIM $\uparrow$ & time (ms) $\downarrow$ & LPIPS $\downarrow$  & L1 $\downarrow$ & FID $\downarrow$ & SSIM $\uparrow$ & time (ms) $\downarrow$\\
\cmidrule{1-20}

    CoordFill~\cite{liu2023coordfill}                     & 0.038  & 0.003  & 12.998 & \textbf{0.972} & 0.039 & 0.021  & 13.011  & 0.977 & 90.3 & 0.044 & 0.004 & 13.2 & \textbf{0.979} & 101.4 & 0.045 & 0.051 & 18.3  & 0.911  & 105\\
    MI-GAN~\cite{Sargsyan_2023_ICCV}                     & \textbf{0.031}  & 0.003  & \textbf{12.932}  & 0.962  & \textbf{0.031}  & 0.003     & \textbf{12.931} & 0.951 & 1200.80 & 0.056  & 0.005  & 21.93  & 0.892 & 3905.4 & 0.121  & 0.112  & 27.34  & 0.832 & 9002.3   \\
    Ours                     & 0.032  & 0.003  & 12.954 & 0.970 & 0.030 & 0.026 & 12.717 & \textbf{0.987} & \textbf{17.59} & \textbf{0.031} & \textbf{0.031} & \textbf{12.5} & 0.975 & \textbf{34.33} & \textbf{0.031} & \textbf{0.029} & \textbf{13.37} & \textbf{0.971} & \textbf{39.43} \\
\bottomrule
\end{tabular}
\end{adjustbox}
\vspace{-2mm}
\caption{{Comparison with state-of-the-art mobile-friendly inpainting methods on different HR inpainting Datasets.}}
\label{table:quantitative}
\end{table*}

\subsection{Attention Upscaling Transfer \& High Frequency Token Mixing}
\label{sec:attention_upscaling}
The attention mechanism of the NeuralPatchMatch module (Sect. \ref{sec:neural_patch_match}) scales quadratically ($O(N^2)$) with the number of patches $N$. Given that $N$ grows linearly with $H$ and $W$, we cannot compute efficiently the attention scores  $\mathbf{A}$ on the high-resolution image $\mathbf{x}$. To meet the latency and memory constraints of real time, we propose a novel post-processing method that can upscale the obtained result ${\hat{\mathbf{x}}}_\text{LR}$ by using the masked inferred attention map $\mathbf{M_T}$ to mix the high-resolution tokens coming directly from image $\mathbf{x}$, in the sense of Equation \eqref{eq:weighted_sum}. This allows us to reduce inference time while  using the full high-quality details from the source image $\mathbf{x}$.
We do so by striding $\mathbf{x}$ proportionally to the learned attention map $\mathbf{A}$. 
In other words, we split the high-resolution image $\mathbf{x}\in\mathbb{R}^{H_{\text{HR}}\times W_{\text{HR}}\times 3}$ into $N$ patches, each of them of size $\frac{H_{HR}}{H}P\times \frac{W_{HR}}{W}P$ pixels.

\begin{figure*}[!ht]
\begin{tabular}{ccccc}
\centering
\hspace*{0.2in}
\begin{tikzpicture}[spy using outlines={thick,red,rectangle,magnification=15,size=3cm,connect spies}]
    \centering
    \node[rectangle,draw,inner sep=0pt] (image) at (0,0){\includegraphics[height=0.2\columnwidth,width=0.2\columnwidth]{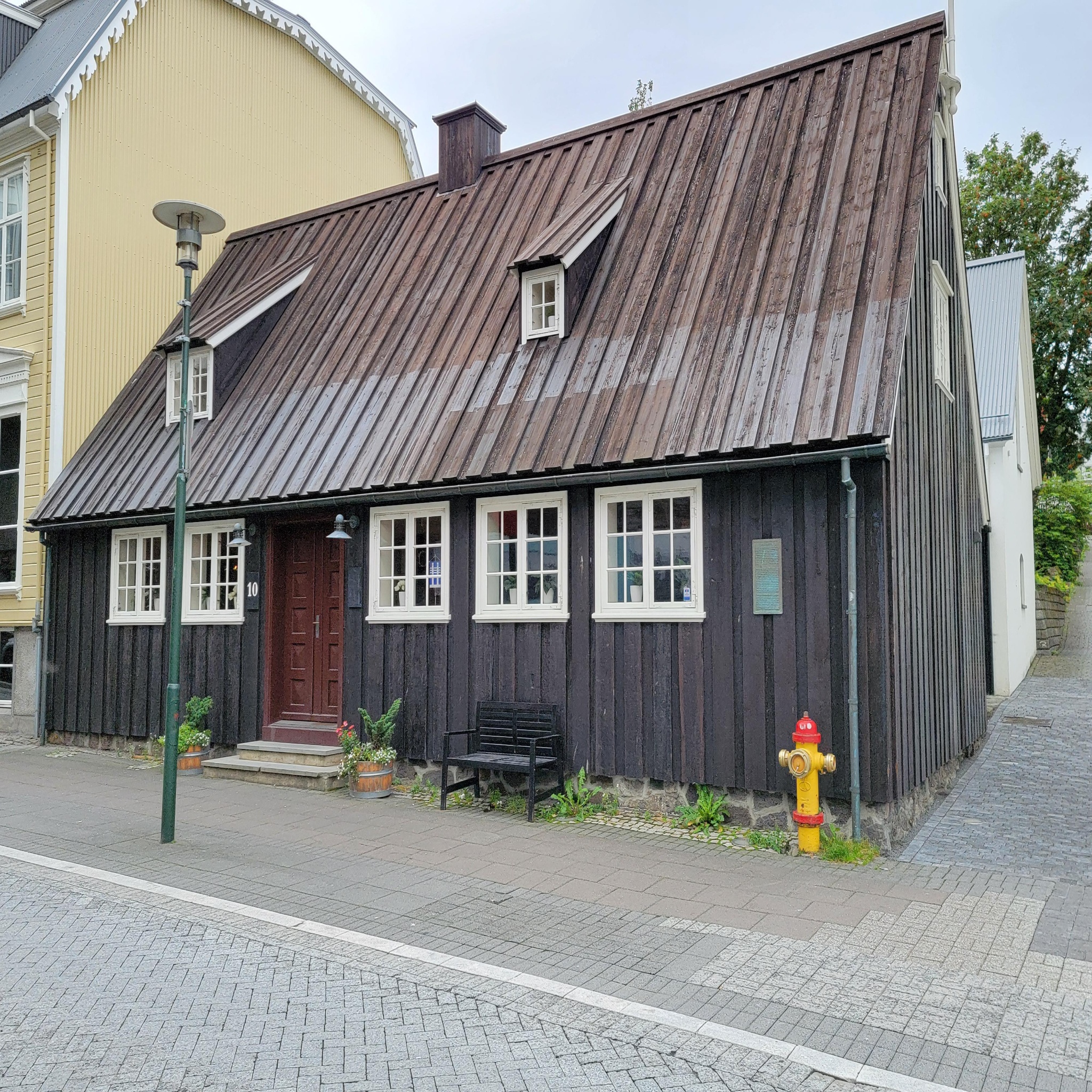}};
    \spy[spy connection path={
}]on (-0.2,0.4) in node at (0.0,0.0);
\end{tikzpicture}&

\hspace*{-0.2in}
\begin{tikzpicture}[spy using outlines={thick,red,rectangle,magnification=15,size=3cm,connect spies}]
    \centering
    \node[rectangle,draw,inner sep=0pt] (image) at (0,0){\includegraphics[height=0.2\columnwidth,width=0.2\columnwidth]{ablation_figures/Ours/DIV8K/20210902/masked_img_20210902_095221_ff_1069_1887_0_0_0.jpg}};
    \spy[spy connection path={
}]on (-0.2,0.4) in node at (0.0,0.0);
\end{tikzpicture}&
\hspace*{-0.2in}
\begin{tikzpicture}[spy using outlines={thick,red,rectangle,magnification=15, size=3cm,connect spies}]
    \centering
    \node[rectangle,draw,inner sep=0pt] (image) at (0,0){\includegraphics[height=0.2\columnwidth,width=0.2\columnwidth]{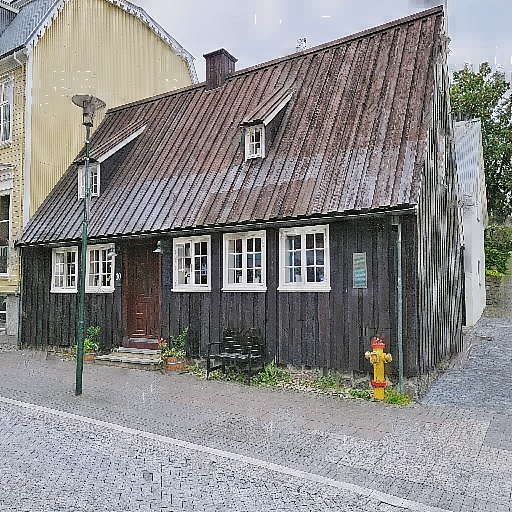}};
    \spy[spy connection path={
}]on (-0.2,0.4) in node at (0.0,0.0);
\end{tikzpicture}&

\hspace*{-0.2in}
\begin{tikzpicture}[spy using outlines={thick,red,rectangle,magnification=15, size=3cm,connect spies}]
    \centering
    \node[rectangle,draw,inner sep=0pt] (image) at (0,0){\includegraphics[height=0.2\columnwidth,width=0.2\columnwidth]{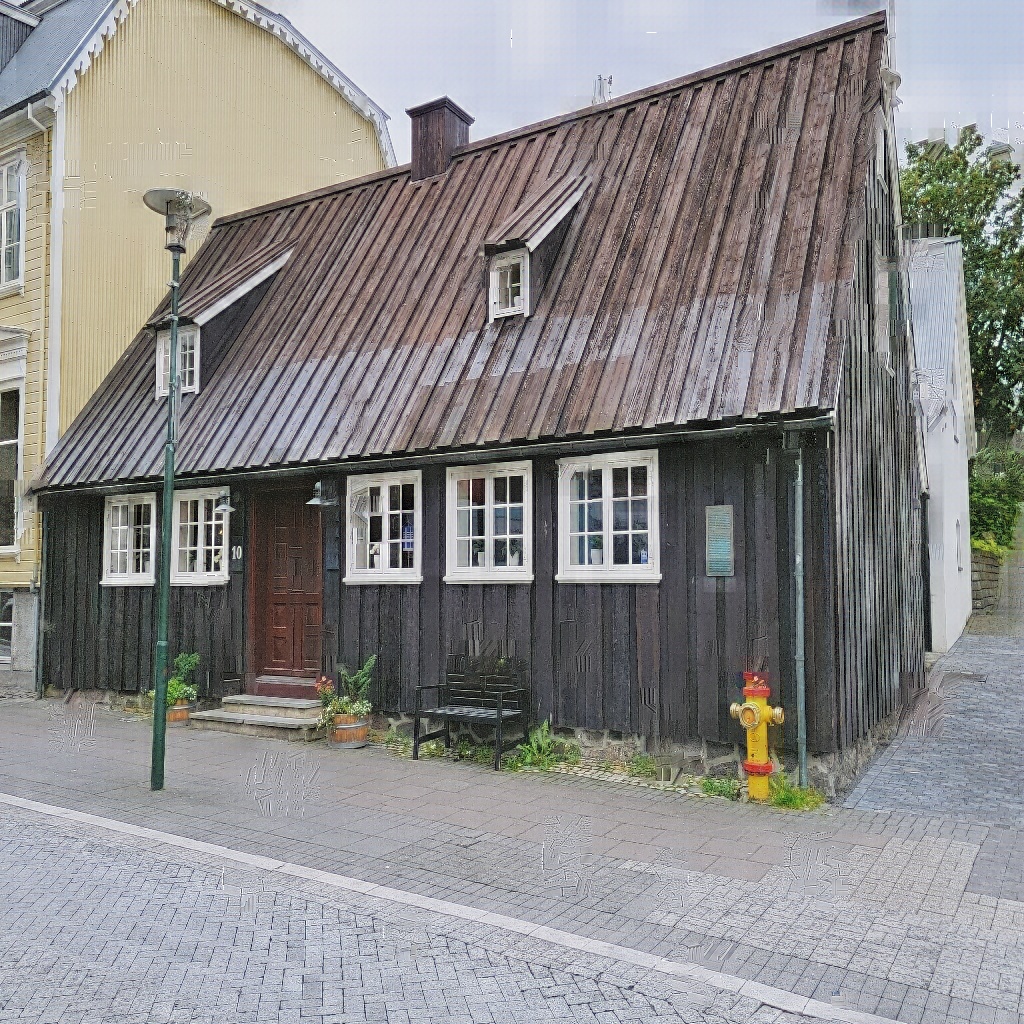}};
    \spy[spy connection path={
}]on (-0.2,0.4) in node at (0.0,0.0);
\end{tikzpicture} &

\hspace*{-0.2in}
\begin{tikzpicture}[spy using outlines={thick,red,rectangle,magnification=15, size=3cm,connect spies}]
    \centering
    \node[rectangle,draw,inner sep=0pt] (image) at (0,0){\includegraphics[height=0.2\columnwidth,width=0.2\columnwidth]{ablation_figures/Ours/DIV8K/20210902/inpainted_high_res_normalized20210902_095221_ff_1069_1887_0_0_0.jpg}};
    \spy[spy connection path={
}]]on (-0.2,0.4) in node at (0.0,0.0);
\end{tikzpicture}%
 \\
\hspace*{0.2in}
\begin{tikzpicture}[spy using outlines={thick,red,rectangle,magnification=15, size=3cm,connect spies}]
    \centering
    \node[rectangle,draw,inner sep=0pt] (image) at (0,0){\includegraphics[height=0.2\columnwidth,width=0.2\columnwidth]{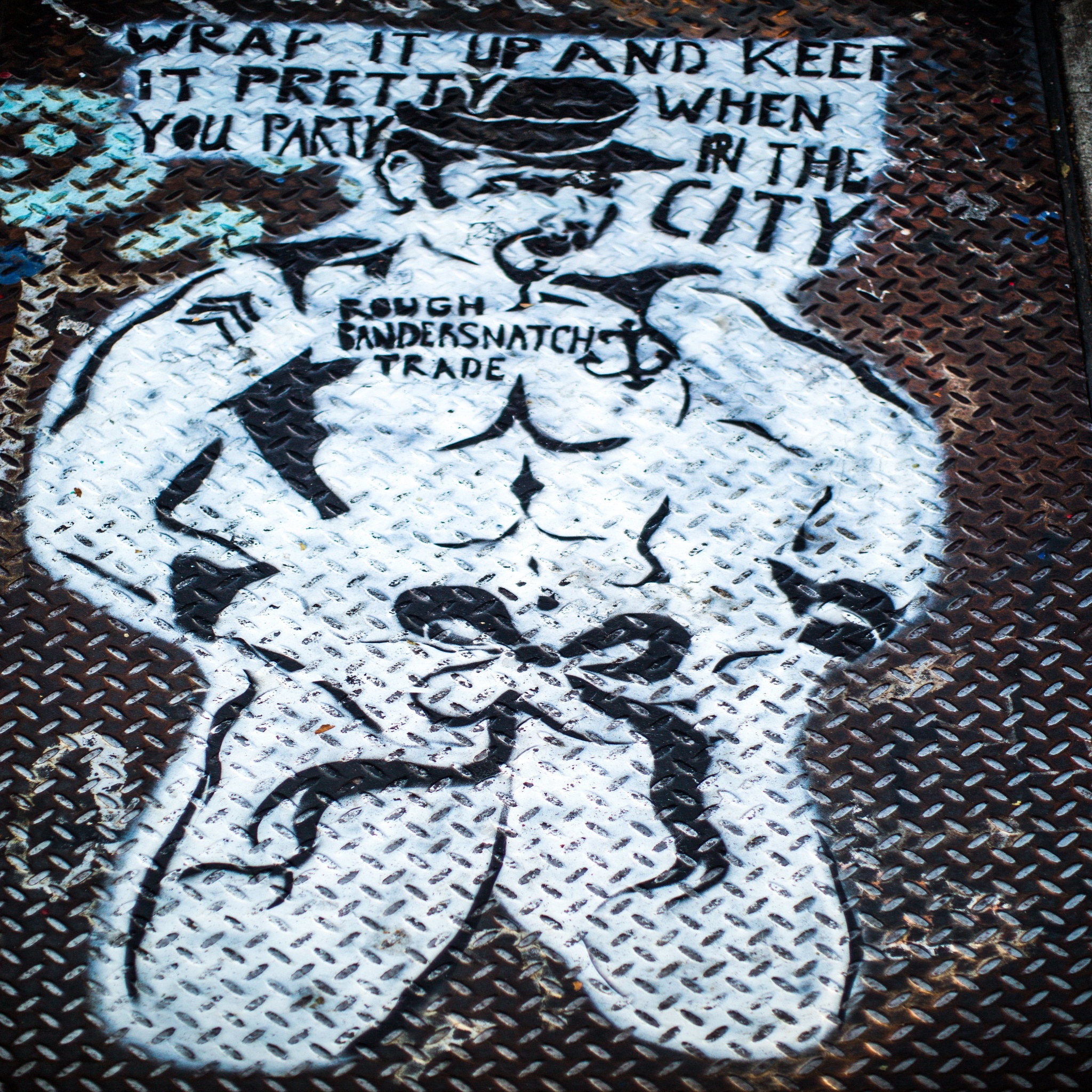}};
    \spy[spy connection path={
}]on (-0.30,0.22) in node at (0.0,0.0);
\end{tikzpicture} & 

\hspace*{-0.2in}
\begin{tikzpicture}[spy using outlines={thick,red,rectangle,magnification=15, size=3cm,connect spies}]
    \centering
    \node[rectangle,draw,inner sep=0pt] (image) at (0,0){\includegraphics[height=0.2\columnwidth,width=0.2\columnwidth]{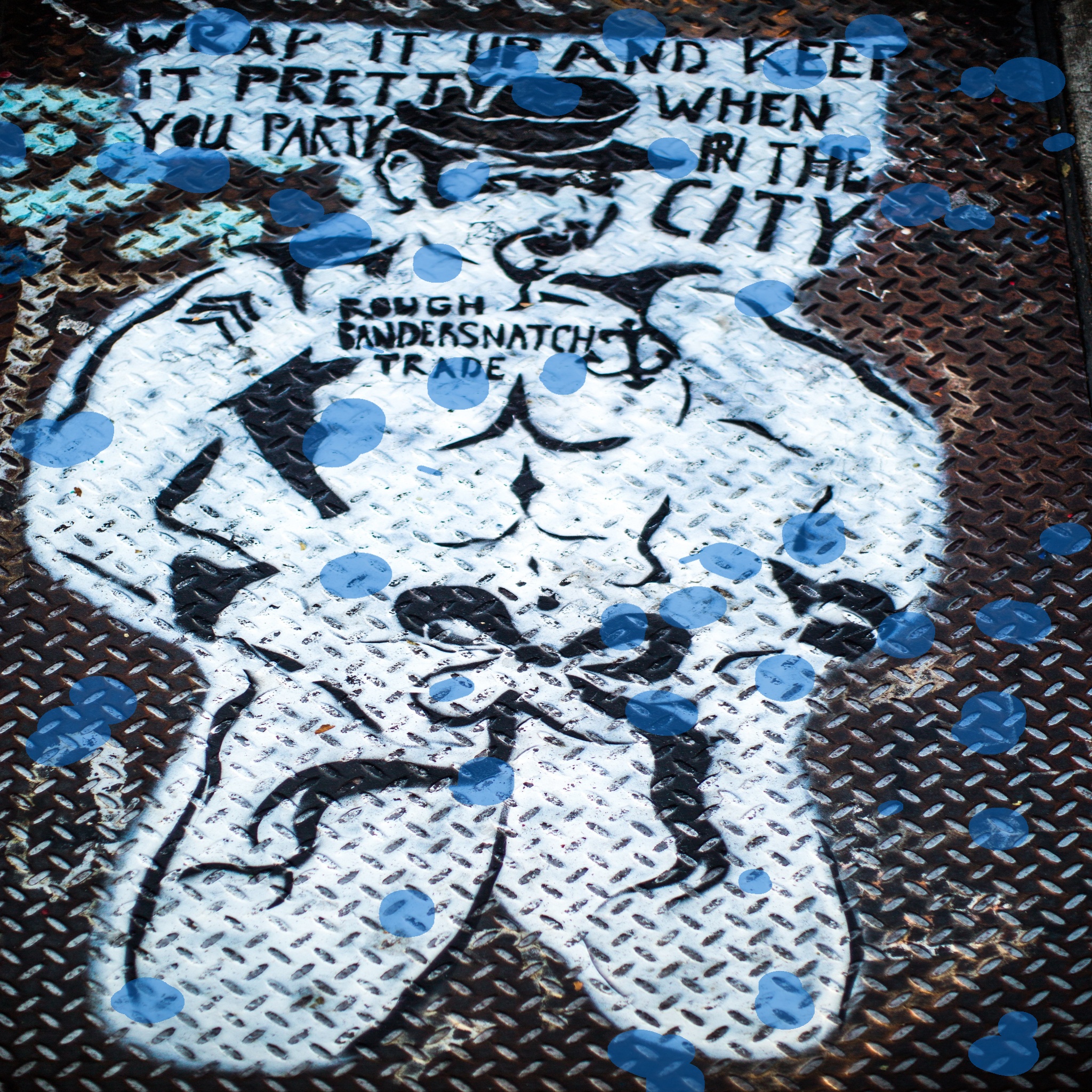}};
    \spy[spy connection path={
}]on (-0.30,0.22) in node at (0.0,0.0);
\end{tikzpicture}&

\hspace*{-0.2in}

\begin{tikzpicture}[spy using outlines={thick,red,rectangle,magnification=15,size=3cm,connect spies}]
\centering 
    \node[rectangle,draw,inner sep=0pt] (image) at (0,0){\includegraphics[height=0.2\columnwidth,width=0.2\columnwidth]{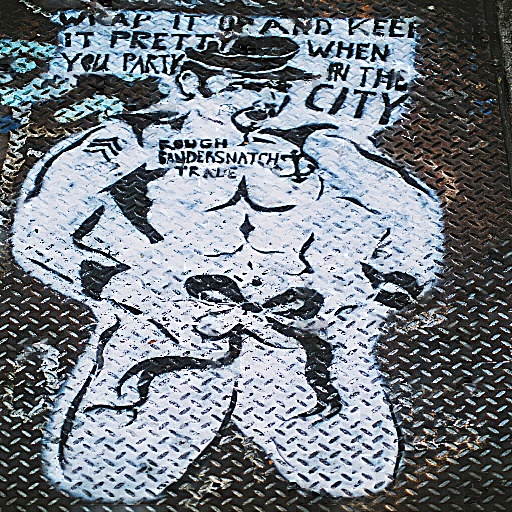}};
    \spy[spy connection path={
}]on (-0.30,0.22) in node at (0.0,0.0);
\end{tikzpicture}&

\hspace*{-0.2in}
\begin{tikzpicture}[spy using outlines={thick,red,rectangle,magnification=15,size=3cm,connect spies}]
\setlength{\tabcolsep}{1pt} 
    \node[rectangle,draw,inner sep=0pt] (image) at (0,0){\includegraphics[height=0.2\columnwidth,width=0.2\columnwidth]{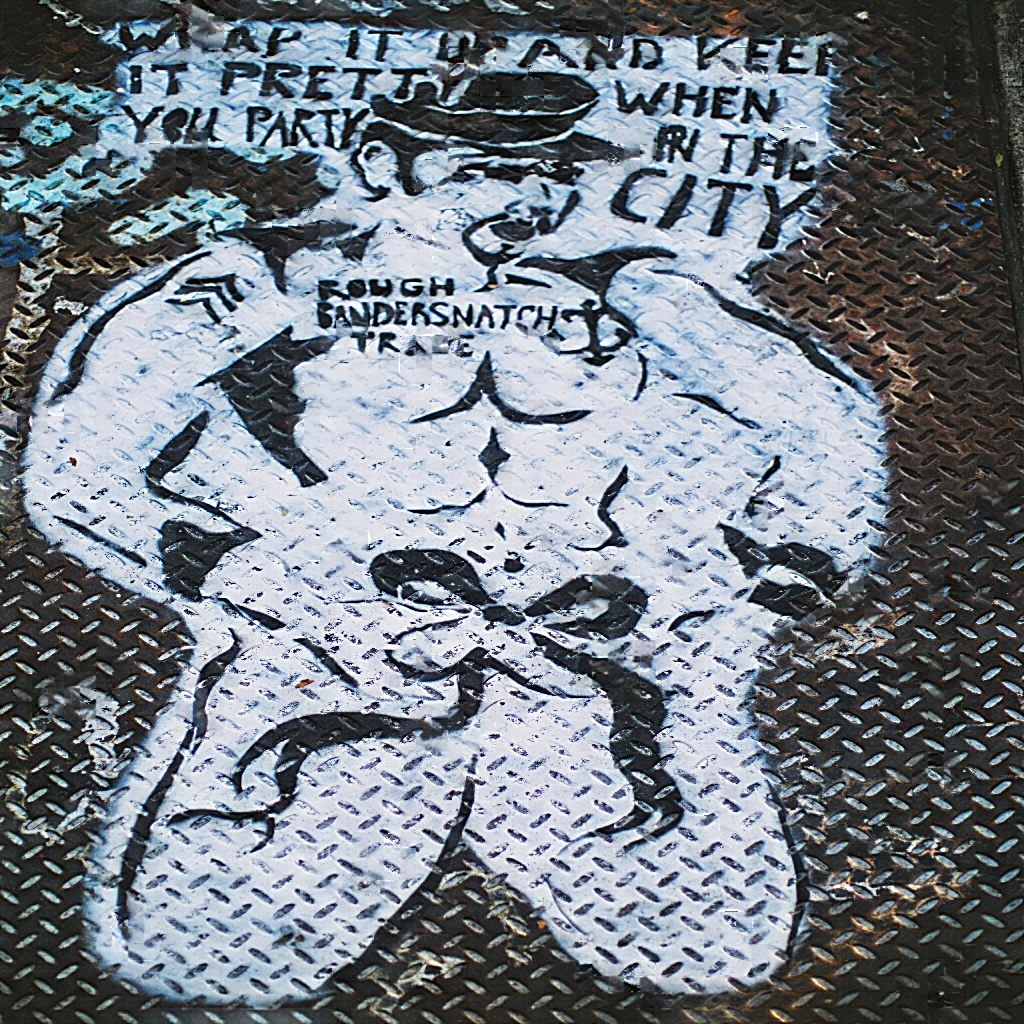}};
    \spy[spy connection path={
}]on (-0.30,0.22) in node at (0.0,0.0);
\end{tikzpicture}&

\hspace*{-0.2in}
\begin{tikzpicture}[spy using outlines={thick,red,rectangle,magnification=15,size=3cm,connect spies}]
\setlength{\tabcolsep}{1pt} 
    \node[rectangle,draw,inner sep=0pt] (image) at (0,0){\includegraphics[height=0.2\columnwidth,width=0.2\columnwidth]{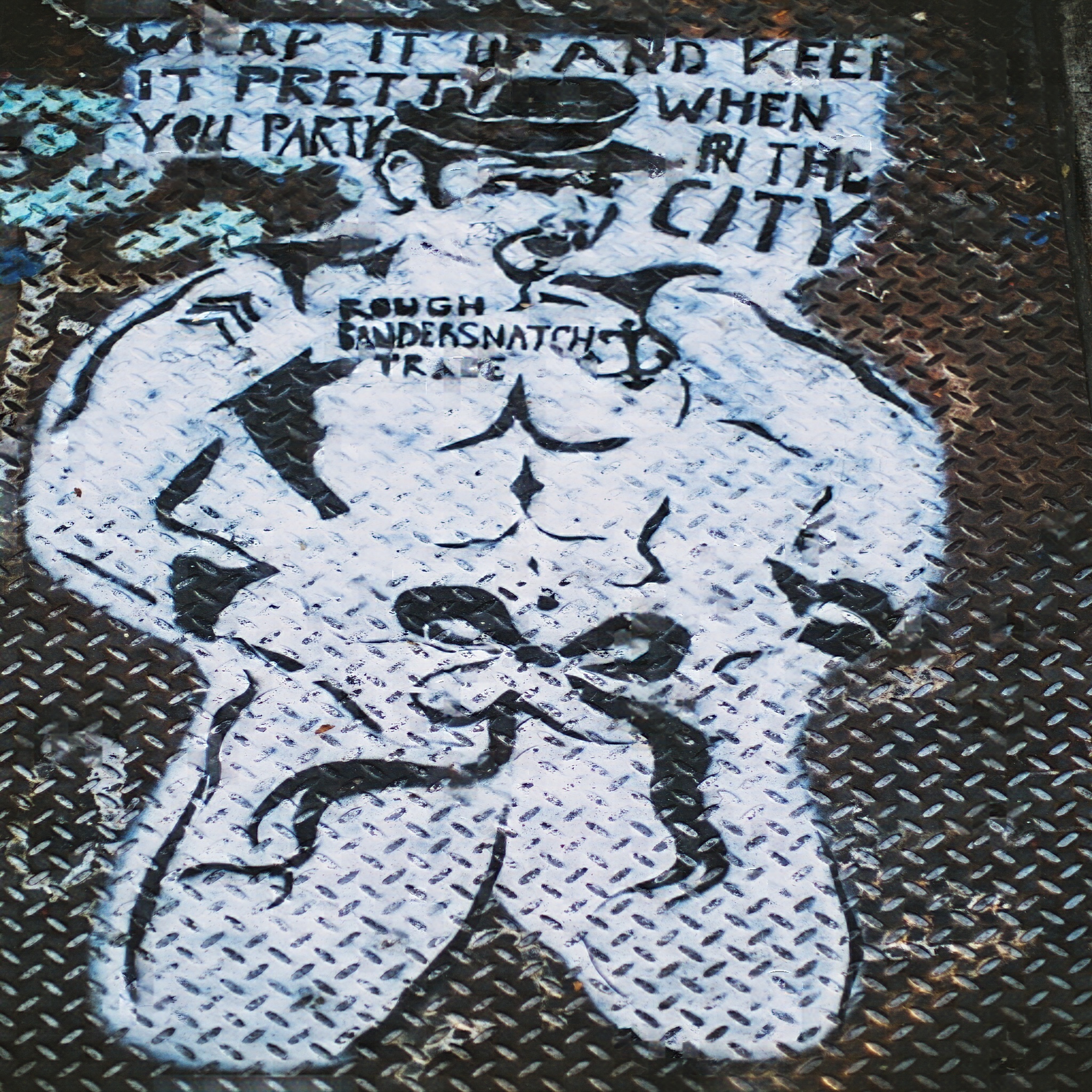}};
    \spy[spy connection path={
}]on (-0.30,0.22) in node at (0.0,0.0);
\end{tikzpicture} \\
 
\hspace*{0.2in}{{Reference HR Image}} & \hspace*{-0.2in}{{Masked Input}} &  \hspace*{-0.2in}{{$512 \times 512$}} & \hspace*{-0.2in}{{$1024 \times 1024$}} & \hspace*{-0.2in}{{$2048 \times 2048$}}
\end{tabular}
\vspace*{-3mm}
\caption{15x zoomed Inpainting results of our proposed method at different higher resolutions. Our method is able to correctly inpaint images at any given resultion, making it suitable for real-world applications.}
\label{fig:importance_of_attention_upscaling}
\end{figure*}

To avoid introducing transitions between patches, which have been carefully suppressed in $\hat{\mathbf{x}}_{LR}$ by the coherence layer, we will use $\mathbf{M_{T}}$ to mix only the high frequencies of the high-resolution tokens~\eqref{eq:high-freq-token}. In this way, we obtain the reconstruction of the high-frequency details~\eqref{eq:hf_weighted_sum} missing in $\hat{\mathbf{x}}_{LR}$ due to the fact that they were originally filtered out by the initial subsampling.
More precisely, let us assume that the initial downsampling has been obtained via a previous proper low-pass filtering with a Gaussian operator $G_{\sigma}$ where the typical deviation $\sigma$ has been carefully chosen to avoid aliasing in the downsampling
~\cite{morel2009asift,otero2014anatomy}. Then, let $G_{\sigma}*\mathbf{X_{HR}}$ be the filtered version of  $\mathbf{X_{HR}}$  and let $\mathbf{p}^{\sigma}_i$ be its patches for $i=1,\dots,N$. Now,  the patches containing the high-frequency components are obtained as:
\begin{equation}
    \mathbf{p}^{HF}_{i} = \mathbf{p}_i - \mathbf{p}^{\sigma}_i.
    \label{eq:high-freq-token}
\end{equation}
for $i=1,\dots,N$. Finally, our reconstructed  high-frequency component is computed as
\begin{equation}
q^{HF}_i= \sum_{j=1}^N M_T(i,j) \mathbf{p}^{HF}_{j} 
 \label{eq:hf_weighted_sum}
\end{equation}
Then, these HF reconstructed patches are rearranged by the Pixel Shuffle step, producing a high-frequency image that is  added to $\hat{\mathbf{x}}_{LR}$ to obtain the final HR inpainting result. This strategy has proven to be effective for very high-resolution inpainting (Fig. \ref{fig:importance_of_attention_upscaling}) while reducing latency cost (Table \ref{table:multiple_res_latency_results}).


\subsection{Model Re-Parametrization}
\label{sec:reparametrization}
Weight re-parametrization is a well known technique \cite{vasu2023fastvit,vasu2023mobileone,Sargsyan_2023_ICCV,zhang2023repnas,ding2019acnet} used in NN literature. Inspired by recent work \cite{chen2022mobile,ding2021repvgg} on inference-time model reparametrization, we adopt this technique on the coarse module to reduce inference latency. Given a block $B_{i}$ on our coarse model   formed by a convolutional layer $C_{i}$ and a batch normalization \cite{ioffe2015batch} layer $b_{i}$, where $C_{i}$ is defined using a kernel of size $S$, number of input channels $C_{\text{in}}$ and output channels $C_{\text{out}}$. We can define its associate weight matrix  as $\mathbf{W} \in \mathbb{R}^{C_{\text{out}} \times C_{\text{in}} \times S \times S}$. For the sake of simplicity, we omit the bias and also the sub-indexes $i$. The batch normalization layer contains a running mean $\mu$, running standard deviation $\sigma$, scale $\gamma$ and bias $\beta$.
To reduce memory bandwidth, we fuse the bach normalization into the convolutional layer and denote it as  $\hat{\mathbf{W}} = \mathbf{W} \frac{\gamma}{\sigma}$. An illustration of the final architecture is displayed on the supplementary material. For the skip connections we follow the same approach as \cite{chen2022mobile,ding2021repvgg}, inside block $B_{i}$ we fuse the batchnorm into a $1 \times 1$ identity convolution. More precisely, we  padd it by $S - 1$ zeros.
\vspace*{-1mm}
\subsection{Latency on Mobile Devices}

Running efficiently NNs on Mobile devices is a non-trivial issue and a general recipe for exporting NNs to different devices does not exist. The edge device ecosystem is highly fragmented \cite{yuan2023rethinking}, a wide landscape where each different chip manufacturer (Qualcomm, Nvidia, Apple)  has its specific accelerators (Hexagon, Tensor Cores, Neural Engine) and model format (DLC, TensorRT, coreml). 
Even though there exists a small subset of NN operations and optimizations that can improve latency on all frameworks such as the ones described in Sect.~\ref{sec:reparametrization},  we must ensure the optimal performance on all popular frameworks in order to create a general mobile-friendly model. Also, many frameworks include pre-defined recipes for common optimizations techniques such as model quantization, palettization or prunning, which improve the inference. In order to prove that the advantage of our method resides on the architecture itself, no post-training optimizations \cite{menghani2021efficient} are applied.

Many methods like \cite{liu2023coordfill,Sargsyan_2023_ICCV} do not design their algorithms mobile-friendly. They include inefficient activations like LeakyRelu \cite{xu2015empirical}, skip connections at inference time that introduce excessive memory access, pooling modules that add excessive synchronization, making them incapable of running in real or semi-real time on most mobile devices.

In addition, previous works only analyzed model characteristics such as the number of parameters or the theoretical float-point operations (FLOPS). Table \ref{table:multiple_res_latency_results} and previous works \cite{ma2018shufflenet,vasu2023mobileone} show that these are indirect metrics that do not correlate with real on-device latency. Device specific factors such as memory cache or registers play an important role. Once again, another factor advocating for the fact that correct testing on multiple edge devices is mandatory in order to prove deployment-ready models.

To analyze the final performance of different mobile inpainting methods, we conduct an extensive benchmark on different edge devices.
%
%
We evaluate our model by exporting to ONNX \cite{ONNX} and Core ML Tools \cite{COREML} format. 
\section{DF8K-Inpainting dataset}

As previously stated, inpainting methods systematically lack evaluation on HR images. To motivate the community to include resolutions above 1K, we release the \emph{DF8K-Inpainting} dataset. This dataset which will be made publicly available upon acceptance is formed by 2850 images. 
It consists of outdoor scenes images, containing a wide variety of entities, from human-made objects to nature landscapes at multiples high-resolutions with no persons in them: 2K, 4K and 8K. This dataset is partitioned into train/test/validation following a 70\%-20\%-10\% split. The fixed masks are provided only for test. The masks are generated similar to \cite{suvorov2022resolution} covering from 30\%-50\% of image extension.  Some examples are displayed in the supplementary material. Let us remark that several efforts from previous works have been done to release HR inpainting datasets \cite{cao2023zits++,Zhang_2022_guided_pm}. Nevertheless, none of them for free-form inpainting masks. The base images are obtained from DF2K original dataset \cite{lim2017enhanced} that merges DIV2K \cite{Timofte_2018_CVPR_Workshops}, FLickr2K \cite{Lim_2017_CVPR_Workshops} and \cite{Zhang_2022_guided_pm} CAFHQ dataset. 
 \section{Experiments}
\subsection{Implementation details}
Our high-resolution inpainting pipeline (displayed in Fig.~\ref{fig:pipeline}) is trained jointly in a single stage. We use Adam optimizer with a learning rate of $10^{-3}$ with warmup and progressive cosine decay. We train all the model versions for 600.000 steps with an effective batch size of 128 in a Nvidia RTX 4090. More information about training recipe can be found in supplementary.



\noindent\textbf{Mask Settings.}
In order to mimic the same masks obtained in real-world mobile inpainting applications, we perform irregular masking of different shapes, similar to \cite{suvorov2022resolution}. At each training iteration, synthetic masks are generated randomly.

\noindent\textbf{Metrics.} We use L1, SSIM \cite{wang2004image}, FID \cite{heusel2017gans} and LPIPS \cite{zhang2018unreasonable} metrics to quantitatively compare the methods.
\subsection{Datasets} In order to test the generalization of our method, we evaluate our proposed pipeline with free-form masks on multiple high-resolution datasets such as CelebA-HQ \cite{karras2019style}, CAF-HQ \cite{Zhang_2022_guided_pm} and our newly introduced DF8K-Inpainting dataset. For all of datasets we create a fix set of testing masks while in training, we randomly generate them at each epoch.

\begin{table*}[!ht]
    \centering
    \begin{adjustbox}{width=1\textwidth}
        \begin{tabular}{lcccccccccc}
        \toprule
    & \multicolumn{2}{c}{Iphone XS} & \multicolumn{2}{c}{Ipad Pro (M2)} & \multicolumn{2}{c}{Ipad 8th generation (A12)} & \multicolumn{2}{c}{Qualcomm Snapdragon 888} & \multicolumn{2}{c}{Jetson Nano} \\
    \cmidrule(lr{0.5em}){2-3} \cmidrule(lr{0.5em}){4-5} \cmidrule(lr{0.5em}){6-7} \cmidrule(lr{0.5em}){8-9} \cmidrule(lr{0.5em}){10-11}
    & Latency (ms) $\downarrow$  &  load (ms) $\downarrow$ & Latency (ms)$\downarrow$  &  load (ms) $\downarrow$ & Latency (ms) $\downarrow$  &  load (ms) $\downarrow$ & Latency (ms) $\downarrow$  &  load (ms) $\downarrow$ & Latency (ms) $\downarrow$ & load (ms) $\downarrow$\\
    
        CoordFill~\cite{liu2023coordfill}   & 2441  & 710 & 90.3 & 126 & 1099 & 1690 &  - & - & 3010.4 & 3010.4 \\
        Mi-GAN~\cite{Sargsyan_2023_ICCV}  & 2300.80   & 814.4  & 1200.80   & 534.4  & 2500 & 1001.3 & 686.17 & 122.3  & 2090 & 3010.4 \\
        \textbf{Ours} & 112.3 &  90.3 & \textbf{17.59}    & \textbf{84.48}  & 74.2 & 209.1 & 101.34 & 118.3 & 1131.3 & 3010.4 \\
        
        \bottomrule
        \end{tabular}
        \end{adjustbox}
         \vspace{-2mm}
        \caption{{\bfseries Experiment: Latency on different devices.}  Inference speed of 3 mobile inpainting networks.}
        \label{table:inferencetimemultipledevices}
        \vspace{-2mm}
\end{table*}
\subsection{Comparison with other Methods}
\noindent\textbf{Competitors:} We compare our proposed model with the few existing mobile state-of-the-art inpainting methods, including CoordFill \cite{liu2023coordfill} and MI-GAN \cite{Sargsyan_2023_ICCV}. Other  well-known inpainting methods such as \cite{suvorov2022resolution, xu2023image, chi2020fast} are omitted as competitors due to its extensive memory footprint, which limits its deployment on edge devices. In order to provide a fair comparison against  fixed-resolution CNN methods such as MI-GAN~\cite{Sargsyan_2023_ICCV}, we infer at trained resolution ($512 \times 512$) and upsample the results to the target resolution.

\noindent\textbf{Latency Benchmarking:} In Table \ref{table:inferencetimemultipledevices} we compare our method against the existing state-of-the-art models for mobile image inpainting. All the methods are exported equally in order to provide a fair comparison. Some models could not be exported to all devices due to unsupported operations such as advanced tensor striding. Our model improves over Coordfill \cite{liu2023coordfill} and Mi-GAN \cite{Sargsyan_2023_ICCV} on FLOPS reduction, model parameters, and most important, latency while obtaining competitive inpainting metrics. The FLOPS count has been computed with \texttt{fvcore} library \cite{fvcore}.

\begin{table}[ht]
    \centering
    \setlength{\tabcolsep}{4pt} 
    \vspace{-3mm}
    \centering
    \resizebox{0.75\columnwidth}{!}{
    \begin{tabular}[t]{lccccc}
    \toprule
&  \multicolumn{1}{c}{1024x1024} &  \multicolumn{1}{c}{2048x2048} &  \multicolumn{1}{c}{Flops@2048} &  \multicolumn{1}{c}{Params} \\
& (ms) &  (ms) & ($\times10^6$) & ($\times10^6$) & \\
    \midrule                                

    CoordFill~\cite{liu2023coordfill}   & 90.3 & 101.4  & 10.73 & 39.6 \\
    Mi-GAN~\cite{Sargsyan_2023_ICCV}  & 1200.80   & 3905.4    & 11.19 & 5.95 \\
    \textbf{Ours}   & \textbf{17.59}    & \textbf{34.33} & \textbf{5.7}  & \textbf{4.3}  \\
    
    \bottomrule
    \end{tabular}}
    \vspace{-1mm}
    \caption{{\bfseries Analysis: Quantitative comparison inference latency.}  Inference speed of different state-of-the-art mobile inpainting networks measured on Ipad Pro at different image resolutions.}
    \label{table:multiple_res_latency_results}
    \vspace{-2mm}
\end{table}

\noindent\textbf{Study on different mobile devices:}
    To correctly evaluate the effectiveness of our model on edge devices, we conduct a large study on the current most popular edge devices platforms. We evaluate on 3 device platforms: Apple, Nvidia and Qualcomm. To provide a fair comparison, we export all methods without quantization and do the inference at fp16.

 \section {Further Analysis \& Ablation Studies}


We conduct an extended ablation study and analysis of our proposal to demonstrate the effectiveness of our methods on two different datasets: CelebHQ \cite{karras2019style} and our newly proposed DF8K-Inpainting dataset.

\noindent\textbf{Coarse Inpainting and the importance of Feature Conditioning:}
We now present an ablation study designed to demonstrate the importance of feature conditioning. 
We find that concatenating the intermediate representations of $f_{\theta}(\cdot)$ improves the overall performance of the model while adding negligible latency over-cost. As anticipated, concatenating the high level features learned on the coarse model helps the NeuralPatchMatch module to learn a better patch affinity (Table \ref{table:ablation_dk_concat}).
We show the performance of inpainting with and without feature conditioning.

It is likely that our NeuralPatchMatch module uses this coarse low-resolution information to retrieve the high-resolution counterpart.

\noindent\textbf{Inpainting Upscaling to higher resolutions via Attention Transfer Module:}
We report the inpainting results at different resolutions obtained via the Attention Transfer Module (Sect. \ref{sec:attention_upscaling}) on Fig.  \ref{fig:importance_of_attention_upscaling}. Our proposed module is able to upscale the learned patch correlation patterns on LR images into HR by utilizing the  high-frequency textures that appear on any given HR image. This multi-resolution behavior is beneficial, providing a resolution-agnostic method that can be deployed on devices with different camera sensors.
\begin{table}[!ht]
    \centering
    \setlength{\tabcolsep}{4pt} 
    \vspace{-3mm}
    \centering
    \resizebox{0.75\columnwidth}{!}{
    \begin{tabular}[t]{lccccc}
    \toprule
    & \multicolumn{1}{c}{Patch Size} & \multicolumn{1}{c}{Parameters} & \multicolumn{1}{c}{Flops $\downarrow$} & \multicolumn{1}{c}{FID $\downarrow$} & \multicolumn{1}{c}{LPIPS $\downarrow$} \\
    & & & ($\times10^6$) & ($\times10^6$) & \\
    
    \cmidrule(lr{0.5em}){1-6}

    & 16 & 15.1 & 3.4 & 27.1 & 0.093 \\
    & 8 & 16.9 & 5.7 & 12.5 & 0.031 \\
    & 32 & 18.3 & 2.1 & 33.5 & 0.134\\
    
    \bottomrule
    \end{tabular}
    }
    \caption{{\bfseries Influence of patch size evaluated on DF8K-Inpainting.}}
    \label{table:ablation_kernel}
    \vspace{-2mm}
\end{table}

\noindent\textbf{Correct Patch Size:}
Table \ref{table:ablation_kernel} shows an evaluation of parameter $P$  on the NeuralPatchMatch \ref{sec:neural_patch_match}.
The patch size is a hyperparameter that plays an important role on the overall image reconstruction. This is likely due to the relation between texture and structure. By increasing the patch size, we can easily inpaint structureless areas, such as open air. However, a smaller patch size allows to reconstruct fine-grained structures and details. Moreover, it  has a high impact on the computational cost, since the computation of the $\mathbf{A}$ depends quadratically on the number of tokens ($O(N^2)$). We choose $P = 8$ as it balances image reconstruction and latency.

\begin{table}[!ht]
    \centering
    \setlength{\tabcolsep}{4pt} 
    \vspace{-5mm}
    \centering
    \resizebox{0.75\columnwidth}{!}{
    \begin{tabular}[t]{lcccccc}
    \toprule
    &  \multicolumn{1}{c}{$d_{K}$} & \multicolumn{1}{c}{Feature Conditioning} & \multicolumn{1}{c}{Parameters} & \multicolumn{1}{c}{Flops $\downarrow$} & \multicolumn{1}{c}{FID $\downarrow$} & \multicolumn{1}{c}{LPIPS $\downarrow$} \\
    & & & ($\times10^6$) & ($\times10^6$) & & \\
    \cmidrule(lr{0.5em}){1-7}

    & 1024     & \checkmark & 16.9 & 4.1 & 13.41 & 0.046 \\
    & 1024     & \xmark & 16.9 & 4.5 & 17.21 & 0.057 \\
    & 2048   & \checkmark & 18.3 & 5.7 & 12.5 & 0.031\\
    & 2048   &  \xmark & 18.3 & 5.3 & 16.43 & 0.052\\
    
    \bottomrule
    \end{tabular}
    }
    \vspace{-2mm}
    \caption{{\bfseries Impact of several model settings evaluated on DF8K-Inpainting.}}
    \label{table:ablation_dk_concat}
    \vspace{-2mm}
\end{table}

\noindent\textbf{Advantages of large embedding dimension:} Table \ref{table:ablation_dk_concat} reports the influence of the embedding dimension. Even though a large embedding dimension adds more capacity to the model to correctly compare the affinity between patches, we can see that it also affects the inference speed. Contrary to popular belief, the inpainting results are hardly improved.


\subsection{Limitations}
Although our method outperforms existing state-of-the-art inpainting solutions, in some challenging situations the model fails to resemble global structure. If the model is not sure enough, it fills the inpainting region by averaging the existing features, creating a blurry effect. In addition, small boundaries between patches can be seen in the  results (Fig. \ref{fig:results_inpainting_FFHQ}). Therefore, always keeping in mind the inference constraints, we will reduce these caveats in future work.

 \section{Conclusions}
In this paper, we propose the first pipeline for high-resolution image inpainting on edge devices. Our method employs local and global methods to inpaint with a coherent structure, while resembling highly detailed textures that appear in the input image. Furthermore, we propose a post-processing step that allows generalization to arbitrary high resolutions. Our method achieves a 100x speedup over other state-of-the-art methods, while achieving similar qualitative and quantitative performance at low resolution and outperforming all previous methods at higher resolutions.

\section{Acknowledgements}

This project is supported by MICINN/FEDER UE project ref. PID2021-127643NB-I00 and Doctorals Industrials ref. DI 2022 075 founded by the Government of Catalonia.

{\small
\bibliographystyle{ieee_fullname}
\bibliography{main}
}

\end{document}